%% file: main.tex
\documentclass{article}

\usepackage{fullpage}

\usepackage[round,comma]{natbib}
\usepackage[utf8]{inputenc} 
\usepackage[T1]{fontenc}    
\usepackage{hyperref}       
\usepackage{url}            
\usepackage{booktabs}       
\usepackage{amsfonts}       
\usepackage{nicefrac}       
\usepackage{microtype}      

\usepackage[colorinlistoftodos, shadow,color=blue!30!white]{todonotes}
\usepackage{amsmath,amsthm,amssymb,amsfonts}
\usepackage{times}

\input{commands}

\newif\ifaistats \aistatsfalse

\allowdisplaybreaks

\title{Model-Free Linear Quadratic Control via Reduction to Expert Prediction}

\author{
Yasin Abbasi-Yadkori \\ Adobe Research 
\and
Nevena Lazi{\'c} \\ Google Brain
\and
Csaba Szepesv{\'a}ri \\ Deepmind
}

\begin{document}

\maketitle

\begin{abstract}
 Model-free approaches for reinforcement learning (RL) and continuous control find policies based only on past states and rewards, without fitting a model of the system dynamics. They are appealing as they are general purpose and easy to implement; however, they also come with fewer theoretical guarantees than model-based RL.  In this work, we present a new model-free algorithm for controlling linear quadratic (LQ) systems, and show that its regret scales as $O(T^{\xi+2/3})$ for any small $\xi>0$ if time horizon satisfies $T>C^{1/\xi}$ for a constant $C$. The algorithm is based on a reduction of control of Markov decision processes to an expert prediction problem. In practice, it corresponds to a variant of policy iteration with forced exploration, where the policy in each phase is greedy with respect to the average of all previous value functions. 
This is the first model-free algorithm for adaptive control of LQ systems that provably achieves sublinear regret and has a polynomial computation cost. Empirically, our algorithm dramatically outperforms standard policy iteration, but performs worse than a model-based approach.
\end{abstract}

\input{introduction.tex}

\input{lq_intro}
\input{estimation_new}

\input{lq_analysis}
\input{experiments.tex}

 \section{DISCUSSION}
 
 The simple formulation and wide practical applicability of LQ control make it an idealized benchmark for studying RL algorithms for continuous-valued states and actions. In this work, we have presented MFLQ, an algorithm for model-free control of LQ systems with an $O(T^{2/3 + \xi})$ regret bound. 
Empirically, MFLQ considerably improves the performance of standard policy iteration in terms of both solution stability and cost, although it is still not cost-competitive with model-based methods. 

Our algorithm is based on a reduction of control of MDPs to an expert prediction problem. In the case of LQ control, the problem structure allows for an efficient implementation and strong theoretical guarantees for a policy iteration algorithm with exploration similar to $\epsilon$-greedy (but performed at a fixed schedule).  While $\epsilon$-greedy is known to be suboptimal in unstructured multi-armed bandit problems~\citep{Langford-Zhang-2007}, it has been shown to achieve near optimal performance in problems with special structure~\citep{abbasi-2009, Rusmevichientong-Tsitsiklis-2010, Bastani-Bayati-2015}, and it is worth considering whether it applies to other structured control problems. 
However, the same approach might not generalize to other domains. For example, Boltzmann exploration may be more appropriate for MDPs with finite states and actions. We leave this issue, as well as the application of $\epsilon$-greedy exploration to other structured control problems, to future work. 

 


\bibliography{biblio}
\newpage
\onecolumn
\appendix
\input{appendix.tex}

\end{document}

%% file: commands.tex
\usepackage{graphicx}

\usepackage{tikz}
\usetikzlibrary{arrows,backgrounds,snakes}

\newcommand{\todonev}[2][]{\todo[size=\scriptsize,color=green!20!white,#1]{Nevena: #2}}
\usepackage{amsmath,abbrevs}
\usepackage{amssymb}

\newtheorem{theorem}{Theorem}[section]
\newtheorem{lemma}[theorem]{Lemma}

\usepackage{algorithm}
\usepackage{algorithmic}

\usepackage{color}
\usepackage{graphicx}
\usepackage{epstopdf}
\usepackage{algorithm,algorithmic}

\usepackage{nicefrac}

\newif\ifcomm
\commtrue

\newif\iflong
\longtrue

\newcounter{assumption}
\renewcommand{\theassumption}{A\arabic{assumption}}

 \usepackage{cleveref}

\newcommand{\ver}[1]{\textsc{v#1}}
\newcommand{\mv}[1]{\textsc{MFLQ#1}}

\newcommand{\estH}{\widehat H}
\newcommand{\estcost}{\overline{c}}
\newcommand{\estG}{\widehat G}

\newcommand{\vect}{\textsc{vec}} 
\newcommand{\mat}{\textsc{mat}}

\newcommand{\norm}[1]{\left\Vert#1\right\Vert}
\newcommand{\abs}[1]{\left\vert#1\right\vert}
\newcommand{\trace}{\mathop{\rm tr}}
\renewcommand{\natural}{\mathbb N}                   
\newcommand{\N}{\mathbb N}                   
\newcommand{\Real}{\mathbb R}                        

\newcommand{\Prob}[1]{{\mathbf P}\left(#1\right)}    
\newcommand{\E}{{\mathbf E}}                         

\newcommand{\argmin}{\mathop{\rm argmin}}

\newcommand{\beq}{\begin{equation}}
\newcommand{\eeq}{\end{equation}}

\ifcomm
   \newcommand\comm[1]{\textcolor{blue}{ #1}}
   \newcommand{\mtodo}[2]{\todo{{\bf #1}: #2}} 
   \def\here#1{{\bf $\langle\langle$#1$\rangle\rangle$}}

\else
   \newcommand\comm[1]{}
   \newcommand{\mtodo}[2]{}
   \def\here#1{}
\fi

\newcommand{\cE}{{\cal E}}
\newcommand{\cZ}{{\cal Z}}
\newcommand{\cX}{{\cal X}}

\newcommand{\cA}{{\cal A}}

\newcommand{\cN}{{\cal N}}

\def\be{\begin{equation}}
\def\ee{\end{equation}}

\newname\controllermixture{{\rm \textsc{MDP-policy-mixture}}}
\newname\stableset{{\rm \textsc{independent-set}}}

%% file: introduction.tex
\ifaistats
\section{INTRODUCTION}
\else
\section{Introduction}
\fi

Reinforcement learning (RL) algorithms have recently shown impressive performance in many challenging decision making problems, including game playing and various robotic tasks. \emph{Model-based} RL approaches estimate a model of the transition dynamics and rely on the model to plan future actions using approximate dynamic programming. 
\emph{Model-free} approaches aim to find an optimal policy without explicitly modeling the system transitions; they either estimate state-action value functions or directly optimize a parameterized policy based only on interactions with the environment.  
Model-free RL is appealing for a number of reasons: 1) it is an ``end-to-end'' approach, directly optimizing the cost function of interest, 2) it avoids the difficulty of modeling and robust planning, and 3) it is easy to implement. However, model-free algorithms also come with fewer theoretical guarantees than their model-based counterparts, 
which presents a considerable obstacle in deploying them in real-world physical systems with safety concerns and the potential for expensive failures. 

In this work, we propose a model-free algorithm for controlling linear quadratic (LQ) systems with theoretical guarantees.
LQ control is one of the most studied problems in control theory \citep{bertsekas1995dynamic}, and it is also widely used in practice. Its simple formulation and tractability given known dynamics make it an appealing benchmark for studying RL algorithms with continuous states and actions. A common way to analyze the performance of sequential decision making algorithms is to use the notion of regret - the difference between the total cost incurred and the cost of the best policy in hindsight \citep{Cesa-Bianchi-Lugosi-2006,hazan2016introduction,shalev2012online}. We show that our model-free LQ control algorithm enjoys a $O(T^{\xi+2/3})$ regret bound. Note that existing regret bounds for LQ systems are only available for model-based approaches. 

Our algorithm is a modified version of policy iteration with exploration similar to $\epsilon$-greedy, but performed at a fixed schedule. Standard policy iteration estimates the value of the current policy in each round, and sets the next policy to be greedy with respect to the most recent value function. By contrast, we use a policy that is greedy with respect to the \emph{average of all past value functions} in each round.  The form of this update is a direct consequence of a reduction of the control of Markov decision processes (MDPs) to expert prediction problems \citep{even2009online}. In this reduction, each prediction loss corresponds to the value function of the most recent policy, and the next policy is the output of the expert algorithm.  The structure of the LQ control problem allows for an easy implementation of this idea: since the value function is quadratic, the average of all previous value functions is also quadratic. 

One major challenge in this work is the finite-time analysis of the value function estimation error. Existing finite-sample results either consider bounded functions or discounted problems, and are not applicable in our setting. Our analysis relies on the contractiveness of stable policies, as well as the fact that our algorithm takes exploratory actions. 
 Another challenge is showing boundedness of the value functions in our iterative scheme, especially considering that the state and action spaces are unbounded. We are able to do so by showing that the policies produced by our algorithm are stable assuming a sufficiently small estimation error.

Our main contribution is a model-free algorithm for adaptive control of linear quadratic systems with strong theoretical guarantees. This is the first such algorithm that provably achieves sublinear regret and has a polynomial computation cost. The only other computationally efficient algorithm with sublinear regret is the model-based approach of \citet{dean2018regret} (which appeared in parallel to this work). 
Previous works have either been restricted to one-dimensional LQ problems~\citep{abeille2017}, or have considered the problem in a Bayesian setting~\citep{ouyang2017}. In addition to theoretical guarantees, we demonstrate empirically that our algorithm leads to significantly more stable policies than standard policy iteration.

\subsection{Related work}


Model-based adaptive control of linear quadratic systems has been studied extensively in control literature.  \emph{Open-loop} strategies identify the system in a dedicated exploration phase. Classical asymptotic results in linear system identification are covered in \citep{ljung1983}; 
an overview of frequency-domain system identification methods is available in \citep{chen2000}, while identification of auto-regressive time series models is covered in \citep{box2015}. 
Non-asymptotic results are limited, and existing studies often require additional stability assumptions on the system \citep{helmicki1991,hardt2016,tu2017coarse}.
\citet{dean2017sample} relate the finite-sample identification error to the smallest eigenvalue of the controllability Gramian. 

Closed-loop model-based strategies update the model online while trying to control the system, and are more akin to standard RL. 
\cite{Fiechter-1997} and \cite{Szita-2007} study model-based algorithms with PAC-bound guarantees for discounted LQ problems. 
Asymptotically efficient algorithms are shown in~\citep{Lai-Wei-1982, Lai-Wei-1987, Chen-Guo-1987, Campi-Kumar-1998, bittanti2006}.   
Multiple approaches \citep{Campi-Kumar-1998,bittanti2006,Abbasi-Yadkori-Szepesvari-2011,ibrahimi2012} have relied on the \emph{optimism in the face of uncertainty} principle. \cite{Abbasi-Yadkori-Szepesvari-2011} show an $O(\sqrt{T})$ finite-time regret bound for an optimistic algorithm that selects the dynamics with the lowest attainable cost from a confidence set; however this strategy is somewhat impractical as finding lowest-cost dynamics is computationally intractable. \cite{abbasi2015,abeille2017,ouyang2017} demonstrate similar regret bounds in the Bayesian and one-dimensional settings using Thompson sampling. \citet{dean2018regret} show an $O(T^{2/3})$ regret bound using robust control synthesis. 

Fewer theoretical results exist for model-free LQ control. The LQ value function can be expressed as a linear function of known features, and is hence amenable to least squares estimation methods. Least squares temporal difference (LSTD) learning has been extensively studied in reinforcement learning, with asymptotic convergence shown by \cite{tsitsiklis1997analysis, Tsitsiklis-VanRoy-1999, yu2009convergence}, and finite-sample analyses given in \cite{antos2008learning, farahmand2016regularized, lazaric2012finite, liu2015finite, liu2012regularized}. 
Most of these methods assume bounded features and rewards, and hence do not apply to the LQ setting. For LQ control, \cite{bradtke1994adaptive} show asymptotic convergence of $Q$-learning to optimum under persistently exciting inputs, and \cite{tu2017least} analyze the finite sample complexity of LSTD for discounted LQ problems. Here we adapt the work of 
 \cite{tu2017least} to analyze the finite sample estimation error in the average-cost setting. Among other model-free LQ methods,  \cite{fazel2018global} analyze policy gradient for deterministic dynamics, and 
\cite{arora2018towards} formulate optimal control as a convex program by relying on a spectral filtering technique for representing linear dynamical systems in a linear basis. 

Relevant model-free methods for finite state-action MDPs include the Delayed Q-learning algorithm of \cite{Strehl-et-al-2006}, which is based on the optimism principle and has a PAC bound in the discounted setting. 
\cite{Osband-etal-2017} propose exploration by randomizing value function parameters, an algorithm that is applicable to large state problems. However the performance guarantees are only shown for finite-state problems.  

Our approach is based on a reduction of the MDP control to an expert prediction problem. The reduction was first proposed by \cite{even2009online} for the online control of finite-state MDPs with changing cost functions. This approach has since been extended to finite MDPs with known dynamics and bandit feedback~\citep{Neu-Gyorgy-Szepesvari-Antos-2014}, LQ tracking with known dynamics~\citep{Abbasi-Yadkori-Bartlett-Kanade-2014}, and linearly solvable MDPs~\citep{Neu-Gomez-2017}.    

\ifaistats
\section{PRELIMINARIES}
\else
\section{Preliminaries}
\fi

We model the interaction between the agent (i.e. the learning algorithm) and the environment as a Markov decision process (MDP). An MDP is a tuple $\langle\cX,\cA,c,P\rangle$, where $\cX\subset\Real^n$ is the state space, $\cA\subset\Real^d$ is the action space, $c:\cX\times\cA\rightarrow \Real$ is a cost function, and $P:\cX\times\cA\rightarrow\Delta_\cX$ is the transition probability distribution that maps each state-action pair to a distribution over states $\Delta_\cX$. 
At each discrete time step $t \in \N$, the agent receives the state of the environment $x_t\in \cX$, chooses an action $a_t\in \cA$ based on $x_t$ and past observations, and suffers a cost $c_t = c(x_t, a_t)$. The environment then transitions to the next state according to $x_{t+1} \sim P(x_t, a_t)$. We assume that the agent does not know $P$, but does know $c$. 
A policy is a mapping $ \pi: \cX \to \cA$ from the current state to an action, or a distribution over actions. Following a policy means that in any round upon receiving state $x$, the action $a$ is chosen according to $\pi(x)$. Let $\mu_\pi(x)$ be the stationary state distribution under policy $\pi$, and let $\lambda_\pi = \E_\mu(c(x, \pi(x))$ be the average cost of policy $\pi$:
\begin{equation*}\label{eq:gain_policy}
\lambda_{\pi} := \lim_{T \rightarrow +\infty } \E_{\pi} \left[\frac{1}{T}\sum_{t = 1}^T c(x_t, a_t) \right] \,,
\end{equation*} 
which does not depend on the initial state in the problems that we consider in this paper. The corresponding bias function, also called value function in this paper, associated with a stationary policy $\pi$ is given by:
\begin{equation*}\label{eq:bias_policy}
V_{\pi}(x) := \lim_{T \rightarrow +\infty} \E_{\pi}\left[ \sum_{t=1}^T(c(x_t, a_t) - \lambda_{\pi}(x_t))\right].
\end{equation*}
The average cost $\lambda_\pi$ and value $V_\pi$ satisfy the following evaluation equation for any state $x\in\cX$,
\begin{equation*}
V_\pi(x) = c(x,\pi(x)) - \lambda_\pi + \E_{x'\sim P(.|x,\pi(x))} (V_\pi(x')).
\end{equation*}
Let $x_t^\pi$ be the state at time step $t$ when policy $\pi$ is followed.  The objective of the agent is to have small regret, defined as
\[
\text{Regret}_T =  \sum_{t=1}^T c(x_t,a_t) - \min_{\pi} \sum_{t=1}^T c(x_t^\pi, \pi(x_t^\pi)) \;.
\]

\subsection{Linear quadratic control}

In a linear quadratic control problem, the state transition dynamics and the cost function are given by
\begin{align*}
x_{t+1} = A x_t + B a_t + w_{t+1}\,, \;\;\;
\ifaistats c_t \else c(x_t,a_t) \fi
= x_t^\top M x_t + a_t^\top N a_t \;.
\end{align*}
The state space is $\cX=\Real^n$ and the action space is $\cA=\Real^d$. We assume the initial state is zero, $x_1=0$. $A$ and $B$ are unknown dynamics matrices of appropriate dimensions, assumed to be controllable\footnote{The linear system is controllable if the matrix $(B ~ AB ~ \cdots ~ A^{n-1}B)$ has full column rank.}. $M$ and $N$ are known positive definite cost matrices. Vectors $w_{t+1}$ correspond to system noise; similarly to previous work, we assume that $w_t$ are drawn i.i.d. from a known Gaussian distribution $\cN(0, W)$. 

In the infinite horizon setting, it is well-known that the optimal policy $\pi_*(x)$ corresponding to the lowest average cost $\lambda_\pi$ 
is given by constant linear state feedback, $\pi_*(x) = - K_* x$. When following any linear feedback policy $\pi(x) = -Kx$, the system states evolve as $x_{t+1} = (A-BK)x_t + w_{t+1}$. A linear policy is called \emph{stable} if $\rho(A-BK) < 1$, where $\rho(\cdot)$ denotes the spectral radius of a matrix. 
It is well-known that the value function $V_\pi$ and state-action value function $Q_\pi$ of any stable linear policy  $\pi(x)= - K x$ are quadratic functions (see e.g. \cite{Abbasi-Yadkori-Bartlett-Kanade-2014}):
\ifaistats
\begin{align*}
Q_\pi(x, a) &= 
\begin{pmatrix} x^\top\, a^\top \end{pmatrix} 
G_\pi 
\begin{pmatrix} x \\  a \end{pmatrix} \\
V_\pi(x) &
= x^\top H_\pi x
= x^\top \begin{pmatrix} I &  -K^\top \end{pmatrix} \,
G_\pi
\begin{pmatrix}  I \\  -K \end{pmatrix} x \,,
\end{align*}
\else
\begin{align*}
Q_\pi(x, a) = 
\begin{pmatrix}
          x^\top\, a^\top
\end{pmatrix} 
G_\pi 
\begin{pmatrix}
           x \\
           a
\end{pmatrix}\,, \;\;\;\;
V_\pi(x) = 
x^\top H_\pi x = x^\top \begin{pmatrix} I &  -K^\top \end{pmatrix} \,
G_\pi
\begin{pmatrix}  I \\  -K \end{pmatrix} x \,,
\end{align*}
\fi
where $H_\pi \succ 0$ and $G_\pi \succ 0$. 
We call $H_\pi$ the value matrix of policy $\pi$.  The matrix $G_\pi$ is the unique solution of the equation
\[
G = 
      \begin{pmatrix} A^\top  \\ B^\top \end{pmatrix}
      \begin{pmatrix} I &  -K^\top \end{pmatrix}
      G      
      \begin{pmatrix} I \\  -K \end{pmatrix}
      \begin{pmatrix} A & B \end{pmatrix}
      +
      \begin{pmatrix}
        M & 0 \\
        0 & N 
       \end{pmatrix}\,. \quad
\]
The greedy policy with respect to $Q_\pi$ is given by
\[
\pi'(x) = \argmin_{a} Q_\pi(x, a) = - G_{\pi,22}^{-1} G_{\pi,21} x = -K x\,.
\] 
Here, $G_{\pi,ij}$ for $i,j\in\{1,2\}$ refers to $(i,j)$'s block of matrix $G_\pi$ where block structure is based on state and action dimensions. The average expected cost of following a linear policy is $\lambda_\pi = \trace(H_\pi W)$.   The stationary state distribution of a stable linear policy is $\mu_\pi(x) = \mathcal{N}(x | 0, \Sigma)$, where $\Sigma$ is the unique solution of the Lyapunov equation
\[
\Sigma = (A-BK) \Sigma (A - BK)^\top + W \,.
\]

%% file: lq_intro.tex
\ifaistats
\section{MODEL-FREE LQ CONTROL}
\else
\section{Model-free control of LQ systems}
\fi

Our model-free linear quadratic control algorithm (\mv{}) is shown in \cref{alg:mflq}, where \ver{1} and \ver{2} indicate different  versions. 
At a high level, \mv{} is a variant of policy iteration with a deterministic exploration schedule. We assume that an initial stable suboptimal policy $\pi_1(x) = - K_1 x$ is given. During phase $i$, we first execute policy $\pi_i$ for a fixed number of rounds, and compute a value function estimate $\widehat V_i$. We then estimate $Q_i$ from $\widehat V_i$ and a dataset  $\cZ = \{(x_t, a_t, x_{t+1})\}$ which includes exploratory actions.
We set $\pi_{i+1}$ to the greedy policy with respect to the average of all previous estimates $\widehat Q_1, ..., \widehat Q_i$. This step is different than standard policy iteration (which only considers the most recent value estimate), and a consequence of using the \textsc{Follow-the-Leader} expert algorithm~\citep{Cesa-Bianchi-Lugosi-2006}. 

The dataset $\cZ$ is generated by executing the policy and  taking a random action every $T_s$ steps. In \mv{v1}, we generate $\cZ$ at the beginning, and reuse it in all phases, while  in \ver{2} we generate a new dataset $\cZ$ in each phase following the execution of the policy. 
While \mv{} as described stores long trajectories in each phase, this requirement can be removed by updating parameters of $V_i$ and $Q_i$ in an online fashion. However, in the case of \mv{v1}, we need to store the dataset $\cZ$ throughout (since it gets reused), so this variant is more memory demanding.

\begin{algorithm}[!t]
{\textsc{MFLQ}} (stable policy $\pi_1$, trajectory length $T$, initial state $x_0$, exploration covariance $\Sigma_a$)
\begin{algorithmic}
\STATE \color{blue} \ver{1}: $S=T^{1/3 - \xi} - 1$, $T_s = const$, $T_v = T^{2/3 + \xi}$
\STATE \ver{1}:
$\cZ = \textsc{CollectData}(\pi_1, T_v, T_s, \Sigma_a)$
\STATE \color{purple} \ver{2}: $S=T^{1/4}$, $T_s=T^{1/4 - \xi}$, $T_v = 0.5 T^{3/4}$
\color{black}
\FOR{$i=1,2,\dots, S$}
\STATE Execute $\pi_i$ for $T_{v}$ rounds and compute  $\widehat V_{i}$ using \eqref{eq:estV}
\STATE \color{purple} \ver{2}: $\cZ = \textsc{CollectData}(\pi_i, T_v, T_s, \Sigma_a)$  \color{black}
\STATE Compute $\widehat Q_{i}$ from $\cZ$ and $\widehat V_{i}$ using \eqref{eq:estQ} 
\STATE $\pi_{i+1}(x)=\argmin_a \sum_{j=1}^i \widehat Q_j (x, a) = -K_{i+1} x$ 
\ENDFOR
\STATE
\end{algorithmic}

\textsc{CollectData} (policy $\pi$, traj. length $\tau$, exploration period $s$,  covariance $\Sigma_a$):
\begin{algorithmic}
\STATE $\cZ = \{\}$
\FOR {$k = 1, 2, \ldots, \lfloor \tau / s \rfloor $}
\STATE Execute the policy $\pi$ for $s-1$ rounds and let $x$ be the final state
\STATE Sample $a \sim \cN(0, \Sigma_a)$, observe next state $x_+$, add $(x, a, x_+)$ to $\cZ$
\ENDFOR
\RETURN $\cZ$
\end{algorithmic}
 \caption{ \textsc{MFLQ}}
\label{alg:mflq}
\end{algorithm}

Assume the initial policy is stable and let $C_1$ be the norm of its value matrix. Our main result are the following two theorems.
\begin{theorem}
\label{thm:ftl-lq2}
For any $\delta,\xi>0$, appropriate constants $C$ and $\overline{C}$, and for $T > \overline{C}^{1/\xi}$, the regret of the \mv{v1} algorithm is bounded as
\[
\text{Regret}_T \le C T^{2/3 + \xi} \log T \;.
\]
\end{theorem}
\begin{theorem}
\label{thm:ftl-lq}
For any $\delta,\xi>0$, appropriate constants $C$ and $\overline{C}$, and for $T > \overline{C}^{1/\xi}$, the regret of the \mv{v2} algorithm is bounded as
\[
\text{Regret}_T \le C T^{3/4 + \xi} \log T \;.
\]
\end{theorem}
The constants $C$ and $\overline{C}$ in the above theorems scale as ${\rm polylog}(n,d,C_1,1/\delta)$. 
To prove the above theorems, we rely on the following regret decomposition. The regret of an algorithm with respect to a fixed policy $\pi$ can be written as
\ifaistats
\[
\text{Regret}_T = \sum_{t=1}^T c_T - \sum_{t=1}^T c_t^{\pi} = \alpha_T + \beta_T + \gamma_T \;,
\]
\[
\alpha_T = \sum_{t=1}^T c_t - \lambda_{\pi_t} ,\; \beta_T =  \sum_{t=1}^T \lambda_{\pi_t} - \lambda_{\pi} , \; \gamma_T =  \sum_{t=1}^T \lambda_{\pi} -  c_t^{\pi}.
\]
\else
\begin{align*}
&\text{Regret}_T = \sum_{t=1}^T c(x_t, a_t) - \sum_{t=1}^T c(x_t^\pi, \pi(x_t^\pi)) = \alpha_T + \beta_T + \gamma_T \;, \\
\text{where}\;\; &\alpha_T = \sum_{t=1}^T (c(x_t,a_t) - \lambda_{\pi_t}) \,,\;\; \beta_T =  \sum_{t=1}^T (\lambda_{\pi_t} - \lambda_{\pi}) \,, \;\; \gamma_T =  \sum_{t=1}^T (\lambda_{\pi} -  c(x_t^\pi, \pi(x_t^\pi))) \;.
\end{align*}
\fi
The terms $\alpha_T$ and $\gamma_T$ represent the difference between instantaneous and average cost of a policy, and can be bounded using mixing properties of policies and MDPs. To bound $\beta_T$, first we can show that (see, e.g. \cite{even2009online})
\[
\lambda_{\pi_t} - \lambda_{\pi} = \E_{x\sim \mu_{\pi}} ( Q_{\pi_t}(x,\pi_t(x)) - Q_{\pi_t}(x, \pi(x)) ) \;.
\]
Let $\widehat Q_{i}$ be an estimate of $Q_{i}$, computed from data at the end of phase $i$. We can write
\begin{align}
\label{eq:beta_decomp}
Q_{i}(x,\pi_i(x)) - Q_{i}(x, \pi(x))  &= \widehat Q_{i}(x,\pi_i(x)) - \widehat Q_{i}(x, \pi(x)) \notag \\
&
+Q_{i}(x,\pi_i(x)) - \widehat Q_{i}(x, \pi_i(x))  \notag \\ 
&
+ \widehat Q_{i}(x,\pi(x)) - Q_{i}(x, \pi(x)) \;.
\end{align}
Since we feed the expert in state $x$ with $\widehat Q_{i}(x,.)$ at the end of each phase, the first term on the RHS can be bounded by the regret bound of the expert algorithm. The remaining terms correspond to the estimation errors. We will show that in the case of linear quadratic control, the value function parameters can be estimated with small error. 
Given sufficiently small estimation errors, we show that all policies remain stable, and hence all value functions remain bounded. 
Given the boundedness and the quadratic form of the value functions, we use existing regret bounds for the FTL strategy to finish the proof.  



%% file: estimation_new.tex
\ifaistats
\section{VALUE FUNCTION ESTIMATION}
\else
\section{Value function estimation}
\fi
\label{sec:TDnew}

\subsection{State value function}
In this section, we study least squares temporal difference (LSTD) estimates of the value matrix $H_\pi$. 
In order to simplify  notation, we will drop $\pi$ subscripts in this section.
\ifaistats
In steady state, with $a_t=\pi(x_t)$, we have the following:
\begin{align*}
V(x_t) &= c(x_t, a_t) + \E(V(x_{t+1}) | x_t, a_t) - \lambda \\
 x_t^\top H x_t &= c(x_t, a_t) + \E(x_{t+1}^\top H x_{t+1} | x_t, a_t) - \trace(WH) \;.
\end{align*}
\else
In steady state, we have the following:
\begin{align*}
V(x_t) &= c(x_t, \pi(x_t)) + \E(V(x_{t+1}) | x_t,\pi(x_t)) - \lambda \\
 x_t^\top H x_t &= c(x_t, \pi(x_t)) + \E(x_{t+1}^\top H x_{t+1} | x_t, \pi(x_t)) - \trace(WH) \;.
\end{align*}
\fi
Let $\vect(A)$ denote the vectorized version of a symmetric matrix $A$, such that $\vect(A_1)^\top \vect(A_2) = \trace(A_1 A_2)$, and let $\phi(x) = \vect(xx^\top)$. We will use the shorthand notation $\phi_t = \phi(x_t)$ and $c_t = c(x_t, a_t)$.  
The vectorized version of the Bellman equation is
\[
 \phi_t^\top \vect(H)  = c_t + \big(\E( \phi_{t+1} | x_t, \pi(x_t)) - \vect(W)\big)^\top H \;.
\]
By multiplying both sides with $\phi_t$ and taking expectations with respect to the steady state distribution,
\[
\E( \phi_t (\phi_{t} - \phi_{t+1} + \vect(W))^\top) \vect(H) = \E( \phi_t c_t ) \;.
\]
We estimate $H$ from data generated by following the policy for $\tau$ rounds.
Let $\Phi$ be a $\tau \times n^2$ matrix whose rows are vectors $\phi_1, ..., \phi_{\tau}$, and similarly let $\Phi_+$ be a matrix whose rows are $\phi_2, ..., \phi_{\tau+1}$. Let ${\bf W}$ be a $\tau  \times n^2$ matrix whose each row is $\vect(W)$. Let ${\bf c} = [c_1, ..., c_{\tau}]^\top$.
The LSTD estimator of $H$ is  given by (see e.g. \citet{Tsitsiklis-VanRoy-1999,yu2009convergence}):
\beq
\label{eq:estV}
\vect(\estH)  = \big(\Phi^\top (\Phi - \Phi_+ + {\bf W}) \big)^{\dagger} \Phi^\top {\bf c} \;,
\eeq
where $(.)^{\dagger}$ denotes the pseudo-inverse.  Given that $H \succ M$, we project our estimate onto the constraint $\estH \succ M$. Note that this step can only decrease the estimation error, since an orthogonal projection onto a closed convex set is contractive.

{\bf Remark}. Since the average cost $\trace(WH)$ cannot be computed from value function parameters $H$ alone, assuming known noise covariance $W$ seems necessary. 
However, if $W$ is unknown in practice, we can use the following estimator instead, which relies on the empirical average cost $\estcost = \frac{1}{\tau} \sum_{t=1}^{\tau} c_t$:
\beq
\label{eq:estV2}
\vect(\widetilde H_{\tau})  = \big(\Phi^\top (\Phi - \Phi_+) \big)^{\dagger} \Phi^\top ( {\bf c} - \estcost {\bf 1}).
\eeq

\begin{lemma}
\label{lem:estv}
\ifaistats
Let $\Sigma_t$ be the state covariance at time step $t$, and let $\Sigma_{\pi}$ be the steady-state covariance of the policy.  Let $\Sigma_V = \Sigma_{\pi} + \Sigma_0$ and let $\Gamma = A-BK$. With probability at least $1 - \delta$, we have
\begin{align}
\| H - \estH\|_F &= \frac{\norm{H}^2}{\lambda_{\min}(M)^2} O\bigg(\tau^{-1/2} \norm{W \estH}_F 
\trace(\Sigma_V) \notag \\
& \norm{\Gamma \Sigma_V^{1/2}} {\rm polylog}(\tau, 1/\delta) \bigg).
\end{align}
\else
Let $\Sigma_t$ be the state covariance at time step $t$, and let $\Sigma_{\pi}$ be the steady-state covariance of the policy.  With probability at least $1 - \delta$, we have
\beq
\| H - \estH\|_F = \frac{\norm{H}^2}{\lambda_{\min}(M)^2} O\bigg(\tau^{-1/2} \norm{W \estH}_F \trace(\Sigma_0 + \Sigma_{\pi})\norm{(A -BK)(\Sigma_0 + \Sigma_{\pi})^{1/2}} {\rm polylog}(\tau, 1/\delta) \bigg).
\eeq
\fi
\end{lemma}

The proof is similar to the analysis of LSTD for the discounted LQ setting by \cite{tu2017least}; however, instead of relying on the contractive property of the discounted Bellman operator, we use the contractiveness of stable policies. 

\begin{proof}
Let $\overline \Phi_+$ be a $\tau \times n^2$ matrix, whose rows correspond to vectors $\E[\phi_2|\phi_1, \pi],  \ldots, \E[\phi_{\tau+1}|\phi_\tau, \pi]$. Let $P_{\Phi} = \Phi(\Phi^\top\Phi)^{-1} \Phi^\top$ be the orthogonal projector onto $\Phi$. 
The value function estimate $\hat{h} = \vect(\estH)$ and the true value function parameters $h =\vect(h)$ satisfy the following:
\begin{align*}
\Phi \hat{h} &= P_{\Phi}({\bf c}  + (\Phi_+ - {\bf W}) \hat h) \\
\Phi h & = {\bf c} + (\overline{\Phi}_+ - {\bf W}) h \;.
\end{align*}
Subtracting the two previous equations, and adding and subtracting $\overline{\Phi}_+ \hat{h}$, we have:
\[
P_{\Phi}( \Phi - \overline{\Phi}_+ + {\bf W})(h - \hat{h}) = P_{\Phi}(\overline \Phi_+ - \Phi_+) \hat{h} \,. 
\]
In Appendix~\ref{app:estv_proof}, we show that the left-hand side can be equivalently be written as 
\[
P_{\Phi}(\Phi - \overline{\Phi}_+ + {\bf W})(h - \widehat{h}) = \Phi (I - \Gamma \otimes \Gamma)^\top \vect(h - \hat h).
\]
Using $\norm{\Phi v} \geq \sqrt{\lambda_{\min}(\Phi^\top \Phi)}\norm{v}$ on the l.h.s., and $P_{\Phi} v \leq \norm{\Phi^\top v} / \sqrt{\lambda_{\min}(\Phi^\top \Phi)}$ on the r.h.s., 
\beq
\label{eq:h_err}
 \norm{ (I - \Gamma \otimes \Gamma)^\top  (h - \hat h)} 
 \leq \frac{ \norm{\Phi^\top (\overline\Phi_+ - \Phi_+) \hat{h}} }
{\lambda_{\min}(\Phi^\top \Phi)}
\eeq
Lemma 4.4 of \cite{tu2017least} shows that for a sufficiently long trajectory, 
$\lambda_{\min}(\Phi^\top \Phi) = O \big(\tau \lambda_{\min}^2(\Sigma_\pi)\big)$. 
Lemma 4.8 of \cite{tu2017least} (adapted to the average-cost setting with noise covariance $W$) shows that for a sufficiently long trajectory, with probability at least $1-\delta$,
\ifaistats
\begin{align}
\label{eq:tu-lemma}
& \norm{\Phi^\top (\overline \Phi_+ - \Phi_+) \hat{h}} 
\leq  \notag \\
& \qquad O \big( \sqrt{\tau}  \trace(\Sigma_V) \norm{ W \estH}_F \norm{\Gamma \Sigma_V^{1/2}}  {\rm polylog}(\tau, 1/\delta) \big).
\end{align}
\else
\begin{align}
\label{eq:tu-lemma}
\norm{\Phi^\top (\overline \Phi_+ - \Phi_+) \hat{h}} 
&\leq   O \bigg( \sqrt{\tau}  \trace(\Sigma_V) \norm{ W \estH}_F \norm{\Gamma \Sigma_V^{1/2}} {\rm polylog}(\tau, 1/\delta) \bigg).
\end{align}
\fi
Here, $\Sigma_V  = \Sigma_0 + \Sigma_\pi$ is a simple upper bound on the state covariances $\Sigma_t$ obtained as follows:
\begin{align*}
\Sigma_{t}  &= \Gamma \Sigma_{t-1} \Gamma^\top + W \\
&= \Gamma^t \Sigma_0 {\Gamma^t}^{\top} + \sum_{k=0}^{t-1} \Gamma^k W {\Gamma^k}^\top  \\
& \prec \Sigma_0  + \sum_{k=0}^{\infty} \Gamma^k W {\Gamma^k}^\top 
 =\Sigma_0 + \Sigma_{\pi}.
 \end{align*}
In Appendix~\ref{app:estv_proof}, we show that $\norm{ (I - \Gamma \otimes \Gamma)^\top  (h - \hat h)}$ is lower-bounded by
$\lambda_{\min}(M)^2 \norm{H}^{-2} \|H - \estH\|_F$. The result follows from applying the bounds to \eqref{eq:h_err} and rearranging terms.
%
\end{proof}

While the error bound depends on $\estH$, for large $\tau$ we have 
\[
\|\estH\|_F - \norm{H}_F \leq \|H - \estH\|_F = c \|\estH\|_F
\]
for some $c < 1$. Therefore $\|\estH\|_F \leq (1 - c)^{-1} \norm{H}_F$. 

\subsection{State-action value function}
\label{sec:estG}

Let $z^\top = (x^\top, a^\top)$, and let $\psi = \vect(z z^\top)$. The state-action value function corresponds to the cost of deviating from the policy, and satisfies
\begin{align*}
Q(x, a) &= z^\top Gz  = c(x, a)  + \E(x_{+}^\top H x_{+} | z) - \trace(HW) \\
\psi^\top \vect(G) &= c(x, a)  + \big(\E(\phi_+ | z) - \vect(W)\big)^\top \vect(H) \;.
\end{align*}
We estimate $G$ based on the above equation, using the value function estimate $\estH$ of the previous section in place of $H$ and randomly sampled actions. 
 Let $\Psi$ be a $\tau \times (n + d)^2$ matrix whose rows are vectors $\psi_1, ..., \psi_{\tau}$, and let ${\bf c} = [c_1, ..., c_{\tau}]^\top$ be the vector of corresponding costs.  
 Let $\Phi_+$ be the $\tau \times n^2$ matrix containing the next-state features after each random action, and let $\overline{\Phi}_+$ be its expectation. 
We estimate $G$ as follows: 
 \beq
 \label{eq:estQ}
 \vect(\estG) = (\Psi^\top \Psi)^{-1} \Psi^\top ({\bf c}  + (\Phi_+ - {\bf W}) \hat{h}),
 \eeq
and additionally project the estimate onto the constraint
$\estG \succeq (\begin{smallmatrix} M & 0 \\ 0 & N \end{smallmatrix})$.

To gather appropriate data, we iteratively (1) execute the policy $\pi$ for $T_s$ iterations in order to get sufficiently close to the steady-state distribution,\footnote{Note that stable linear systems mix exponentially fast; see \cite{tu2017least} for details.} (2) sample a random action $a \sim \cN(0, \Sigma_a)$, (3) observe the cost $c$ and next state $x_+$, and (4) add the tuple $(x, a, x_+)$ to our dataset $\cZ$.  We collect $\tau = 0.5 T^{1/2 + \xi}$ such tuples in each phase of \mv{v2}, and $\tau =  T^{2/3 + \xi}$ such tuples in the first phase of \mv{v1}.

\begin{lemma}
\label{lem:est-error}
Let $\Sigma_{G,\pi} = A \Sigma_\pi A^\top + B \Sigma_a B^\top$. 
With probability at least $1-\delta_1$, we have
\ifaistats
\begin{align}
& \norm{G - \estG}_F  =   O \bigg( \trace(\Sigma_{G,\pi})  \norm{H - \estH}  \notag \\
 &\qquad +  {\tau}^{-1/2} \norm{ W \widehat{H}}_F \trace (\Sigma_\pi+ \Sigma_a) \norm{\Sigma_{G,\pi}}_F \notag \\
 & \qquad \quad \times {\rm polylog}(n^2, \frac{1}{\delta_1}, \tau) \bigg).
\end{align}
\else
\begin{align}
\norm{G - \estG}_F  
= & \; O \bigg( \trace(\Sigma_{G,\pi})  \norm{H - \estH}  \notag \\
 &+  {\tau}^{-1/2} \norm{ W \widehat{H}}_F (\trace (\Sigma_\pi) + \trace (\Sigma_a)) \norm{\Sigma_{G,\pi}}_F  {\rm polylog}(n^2, 1/\delta_1, \tau) \bigg).
\end{align}
\fi
\end{lemma}
The proof is given in Appendix~\ref{app:estg_proof} and similar to that of Lemma~\ref{lem:estv}. One difference is that we now require a lower bound on $\lambda_{\min}(\Psi^\top \Psi)$, where $\Psi^\top\Psi$ is a function of both states and actions. Since actions may lie in a subspace of $\cA$ when following a linear policy, we estimate $G$ using only the exploratory dataset $\cZ$. Another difference is that we rely on $\widehat H$, so the error includes a $\|H - \widehat H\|$ term.

Let $\Sigma_{\max}$ be an upper bound on the state covariance matrices in the p.s.d. sense, and let $C_H$ be an upper bound on $\norm{H_i}$ (see Appendix~\ref{sec:cov_bound} and Lemma~\ref{lem:greedy-stable} for concrete values for these bounds). 
By a union bound, with probability at least $1 - S (\delta + \delta_1)$, the estimation error in any phase $i$ of \mv{} with $\tau_v$ steps of value estimation and $\tau_q$ random actions is bounded as
\ifaistats
\begin{align}
\label{eq:errG}
& \norm{G_i - \widehat{G}_i}_F \leq \varepsilon_1 :=  \notag \\
& C_h \tau_v^{-1/2} \trace(\Sigma_{G, \max}) \frac{C_H^3 \norm{W}_F \trace(\Sigma_{\max})}{\lambda_{\min}(M)}  \norm{\Gamma \Sigma_{\max}^{1/2}}  \notag \\
 &+ C_{g} {\tau_q}^{-1/2} C_H \norm{ W}_F (\trace (\Sigma_{\max}) + \trace (\Sigma_a)) \norm{\Sigma_{G, \max}}_F 
\end{align}
\else
\begin{align}
\label{eq:errG}
\norm{G_i - \widehat{G}_i}_F \leq \varepsilon_1 := & \;
C_h \tau_v^{-1/2} \trace(\Sigma_{G, \max}) \frac{C_H^3}{\lambda_{\min}(M)}  \norm{W}_F \trace(\Sigma_{\max}) \norm{(A-BK) \Sigma_{\max}^{1/2}} {\rm polylog}(\tau_v, 1/\delta) \notag \\
 &+ C_{g} {\tau_q}^{-1/2} C_H \norm{ W}_F (\trace (\Sigma_{\max}) + \trace (\Sigma_a)) \norm{\Sigma_{G, \max}}_F {\rm polylog}(n^2, 1/\delta_1, \tau_q)
\end{align}
\fi
for appropriate constants $C_h$ and $C_g$.

%% file: lq_analysis.tex
\ifaistats
\section{ANALYSIS OF \mv{}}
\else
\section{Analysis of the \mv{} algorithm}
\fi
\label{sec:lq_analysis}

In this section, we first show that given sufficiently small estimation errors, all policies produced by the \mv{} algorithm remain stable. Consequently the value matrices, states, and actions remain bounded. We then bound the terms $\alpha_T$, $\beta_T$, and $\gamma_T$ to show the main result. 
For simplicity, we will assume that $M \succ I$ and $N \succ I$ for the rest of this section; we can always rescale $M$ and $N$ so that this holds true without loss of generality. We analyze \mv{v2}; the analysis of \mv{v1} is similar and obtained by a different choice of constants. 

By assumption, $K_1$ is bounded and stable. By the arguments in Section~\ref{sec:TDnew}, the estimation error in the first phase can be made small for sufficiently long phases. In particular, we assume that estimation error in each phase is bounded by $\varepsilon_1$ as in Equation~\eqref{eq:errG} and that $\varepsilon_1$ satisfies 
\beq
\label{eq:eps1}
 \varepsilon_1 < \big(12 C_1 (\sqrt{n} + C_K \sqrt{d})^2 S \big)^{-1}\;.
\eeq
Here $C_1>1$ is an upper bound on $\norm{H_1}$, and  $C_K = 2( 3 C_1 \norm{B}\norm{A} + 1)$. Since we take $S^2 T^{\xi}$ random actions in the first phase, the error factor $S^{-1}$ is valid as long as $T > \overline{C}^{1/\xi}$ for a constant $\overline{C}$ that can be derived from \eqref{eq:errG} and \eqref{eq:eps1}. 
We prove the following lemma in Appendix~\ref{sec:stable-proof}.
\begin{lemma}
\label{lem:greedy-stable}
Let $\{K_i\}_{i=2}^S$ be the sequence of policies produced by the MFLQ algorithm. For all $i\in [S]$, $\norm{H_i} \le C_i < C_H := 3C_1$, $K_i$ is stable,  $\norm{K_i} \le C_K$, and for all $k\in\natural$,
\ifaistats
\begin{align*}
\norm{(A-BK_i)^k}  & \leq \sqrt{C_{i-1}}(1 - (6C_1)^{-1})^{k/2} \\
& \leq \sqrt{C_H}(1 - (2C_H)^{-1})^{k/2}\;.
\end{align*}
\else
\[
\norm{(A-BK_i)^k}  \leq \sqrt{C_{i-1}}(1 - (6C_1)^{-1})^{k/2}
\leq \sqrt{C_H}(1 - (2C_H)^{-1})^{k/2}\;. 
\]
\fi
\end{lemma}
To prove the lemma, we first show that the value matrices $H_j$ and closed-loop matrices $\Gamma_j = A - BK_j$ satisfy
\begin{align}
\label{eq:stab_ineq2}
x^\top \big( \Gamma_{j+1}^\top H_j \Gamma_{j+1} \big)x  +  \varepsilon_2 \leq x^\top H_j x
\end{align}
where
\[
\varepsilon_2 = 
x^\top (M + K_{j+1}^\top N K_{j+1})x  - \varepsilon_1 {\bf 1}^\top (|X_{K_j}| + |X_{K_{j+1}}|){\bf 1}\,,
\]
\[X_K = \begin{bmatrix}I \\ -K \end{bmatrix} xx^\top \begin{bmatrix}I & -K^\top \end{bmatrix} ,\] 
and $|X_K|$ is the matrix obtained from $X_K$ by taking the absolute value of each entry. 
If the estimation error $\varepsilon_1$ is small enough so that $\varepsilon_2 > 0$ for any unit-norm $x$ and all policies, then $H_j \succ \Gamma_{j+1}^\top H_j \Gamma_{j+1}$ and $K_{j+1}$ is stable by a Lyapunov theorem. Since $K_1$ is stable and $H_1$ bounded, all policies remain stable. If estimation errors are bounded as in \eqref{eq:eps1}, we can show that the policies and value matrices are bounded as in Lemma~\ref{lem:greedy-stable}.
 
 Let $\cE$ denote the event that all errors are bounded as in Equation~\eqref{eq:eps1}. We bound the terms $\alpha_T$, $\beta_T$, and $\gamma_T$ for \mv{v2} next. 
\begin{lemma}
\label{lem:bound-beta}
Under the event $\cE$, for appropriate constants $C'$, $D'$, and $D''$, 
\[
\beta_T \le C' T^{3/4} \log (T/\delta), \; 
\alpha_T \le D' T^{3/4 + \xi}, 
\; \ifaistats \else {\rm and} \fi \;
\gamma_T \le D'' T^{1/2} \;.
\]
\end{lemma}
The proof is given in Appendix~\ref{sec:regret_bound}.
For $\beta_T$, we rely on the decomposition shown in Equation \eqref{eq:beta_decomp}. Because we execute $S$ policies, each for $\tau = T / S$ rounds (where $S = T^{1/3 - \xi}$ and $\tau = T^{1/4}$ for \mv{v1} and \mv{v2}, respectively),
\begin{align*}
\beta_T &= \sum_{i=1}^S \tau \E ( Q_{i}(x,\pi_i(x)) - Q_{i}(x, \pi(x)) ) \\
&= \tau \sum_{i=1}^S \bigg( \E ( \widehat Q_{i}(x,\pi_i(x)) - \widehat Q_{i}(x, \pi(x)) )\\ 
&\qquad \qquad  +  \E ( Q_{i}(x,\pi_i(x)) - \widehat Q_{i}(x, \pi_i(x)) ) \\
&\qquad \qquad  +  \E  ( \widehat Q_{i}(x,\pi(x)) - Q_{i}(x, \pi(x)) ) \bigg) 
\end{align*}
where the expectations are w.r.t. $x \sim \mu_\pi$. We bound the first term using the FTL regret bound of \citep{Cesa-Bianchi-Lugosi-2006} (Theorem 3.1), by showing that the theorem conditions hold for the loss function $ f_i(K) = \E_{x \sim \mu_{\pi}}( \widehat Q_{i}(x, Kx))$. We bound the second and third term (corresponding to estimation errors) using Lemma~\ref{lem:est-error}. This results in the following bound on $\beta_T$ for constants $C'$ and $C''$:
\[
\beta_T \le T / S C' (1 + \log S) +  C' \sqrt{ST} \log T  \;.
\]

To bound $\gamma_T = \sum_{t=1}^T \lambda_{\pi} -  c(x_t, \pi(x_t))$, we first decompose the cost terms as follows. 
Let $\Sigma_\pi$ be the steady-state covariance, and let $\Sigma_t$ be the covariance of $x_t$. Let $D_t = \Sigma_t^{1/2}(M + K^\top N K)\Sigma_t^{1/2}$ and  $\lambda_t = \trace(D_t)$. We have
\ifaistats
\begin{align*}
\lambda_\pi - c(x_t, \pi(x_t) ) 
& = \lambda_\pi - \lambda_t + \lambda_t - c(x_t^\pi, \pi(x_t^\pi)) \\
&= \trace((\Sigma_\pi - \Sigma_t)(M + K^\top N K))   \\
&\quad +  \trace(D_t) - u_t^\top D_t u_t
\end{align*}
\else
\begin{align}
\lambda_\pi - c(x_t, \pi(x_t)) &= \lambda_\pi - \lambda_t + \lambda_t - c(x_t^\pi, \pi(x_t^\pi)) \\
&= \trace((\Sigma_\pi - \Sigma_t)(M + K^\top N K))  + \big( \trace(D_t) - u_t^\top D_t u_t\big)
\end{align}
\fi
where $u_t \sim \mathcal{N}(0, I_n)$.  We show that the second term $\sum_{t=1}^T \trace(D_t) - u_t^\top D_t u_t$ scales as $\sqrt{T}$ with high probability using the Hanson-Wright inequality. The first term can be bounded by $\trace(H_\pi)\trace(\Sigma_\pi)$ as follows.  Note that 
\[
\Sigma_\pi - \Sigma_t =  \Gamma (\Sigma_\pi - \Sigma_{t-1}) \Gamma^\top =  \Gamma^t(\Sigma_\pi - \Sigma_0)(\Gamma^t)^\top.
\]
Hence we have
\ifaistats
\begin{align}
& \sum_{t=0}^T \trace((M + K^\top N K) (\Sigma_\pi - \Sigma_t))  \notag \\
&\qquad =  \sum_{t=0}^T  \trace \big( (M + K^\top N K) \Gamma^t(\Sigma_\pi - \Sigma_0)(\Gamma^t)^\top \big) \notag \\
&\qquad \leq \trace(\Sigma_\pi - \Sigma_0) \trace \bigg( \sum_{t=0}^\infty (\Gamma^t)^\top (M + K^\top N K) \Gamma^t \bigg) \notag \\
& \qquad = \trace(\Sigma_\pi - \Sigma_0) \trace( H_\pi ) \;.
\label{eq:sigma_bound}
\end{align}
\else
\begin{align}
\sum_{t=0}^T \trace((M + K^\top N K) (\Sigma_\pi - \Sigma_t)) 
&=  \sum_{t=0}^T  \trace \big( (M + K^\top N K) \Gamma^t(\Sigma_\pi - \Sigma_0)(\Gamma^t)^\top \big) \\
&\leq \trace(\Sigma_\pi - \Sigma_0) \trace \bigg( \sum_{t=0}^\infty (\Gamma^t)^\top (M + K^\top N K) \Gamma^t \bigg)\\
&= \trace(\Sigma_\pi - \Sigma_0) \trace( H_\pi ) \;.
\label{eq:sigma_bound}
\end{align}
\fi
The bound on  $\alpha_T = \sum_{t=1}^T c(x_t,a_t) - \lambda_{\pi_t}$ is similar; however, in addition to bounding the cost of following a policy, we need to account for the cost of random actions, and the changes in state covariance due to random actions and policy switches.

Theorem~\ref{thm:ftl-lq} is a consequence of Lemma~\ref{lem:bound-beta}. 
The proof of Theorem~\ref{thm:ftl-lq2} is similar and is obtained by different choice of constants. 

%% file: experiments.tex
\ifaistats
\section{EXPERIMENTS}
\else
\section{Experiments}
\fi

 \begin{figure*}[!t]
 \centering
 \includegraphics[width=0.95\linewidth, trim=0cm 0cm 0.5cm 0cm, clip]{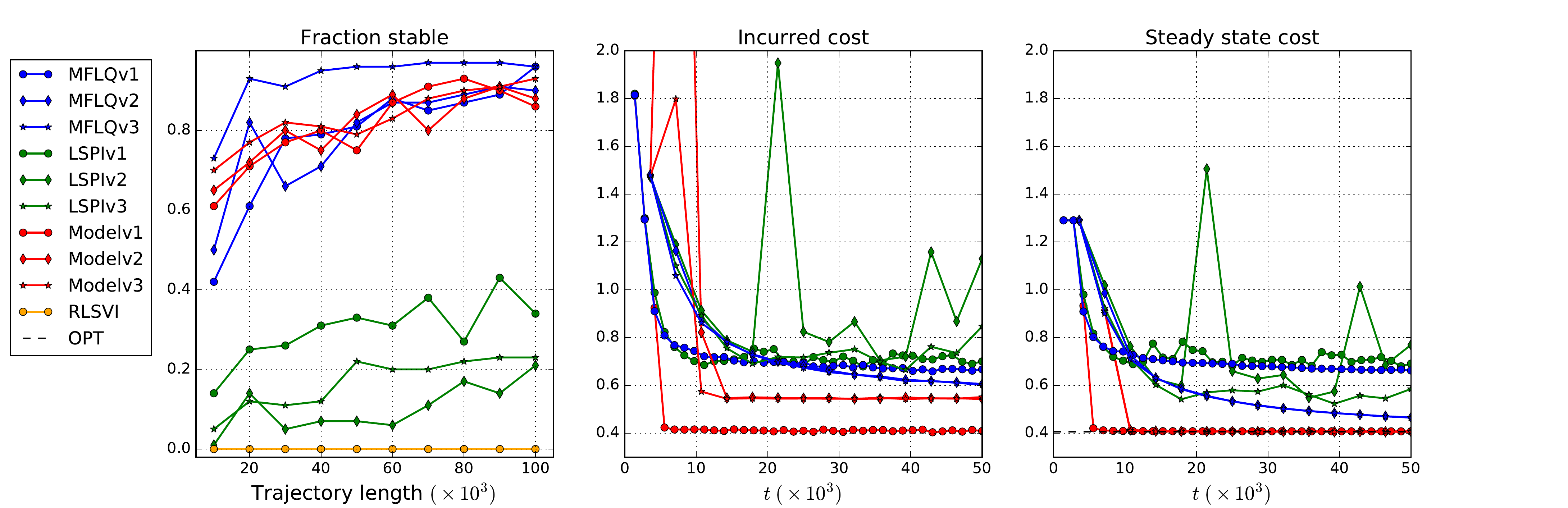}
\includegraphics[width=0.95\linewidth, trim=0cm 0cm 0.5cm 0cm, clip]{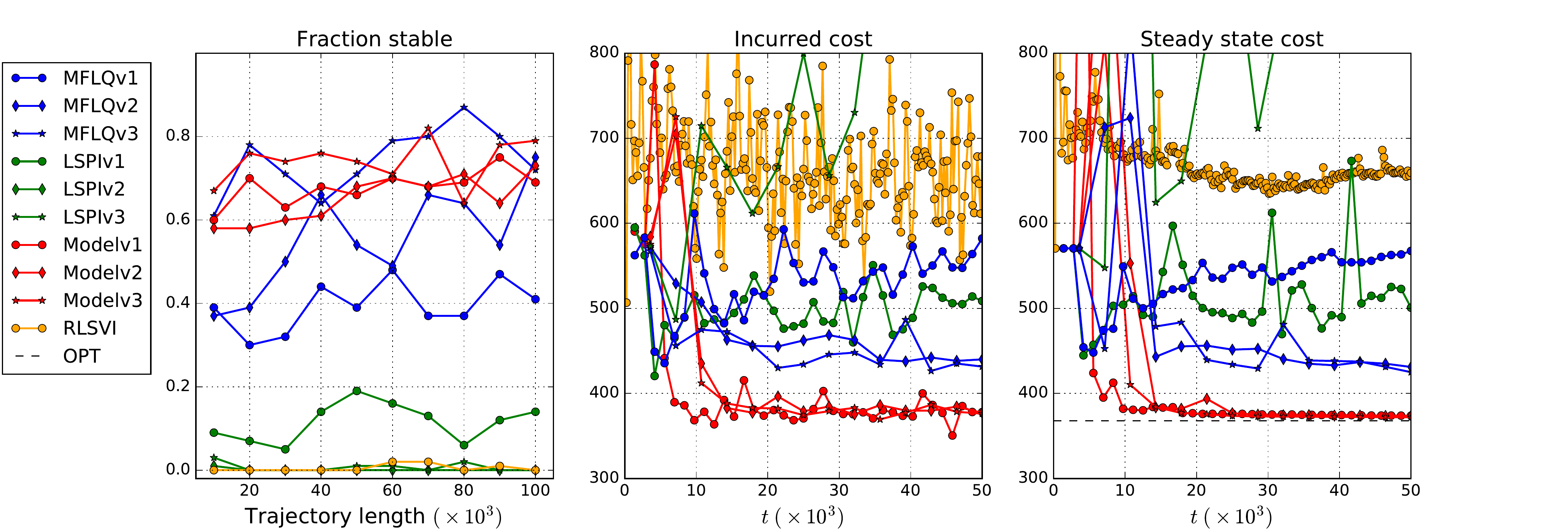}
\caption{Top row: experimental evaluation on the dynamics of \cite{dean2017sample}. Bottom row: experimental evaluation on \cite{lewis2012optimal}, Example 11.5.1. 
}
\label{fig:experiments}
\end{figure*}

We evaluate our algorithm on two LQ problem instances: (1) the system studied in \cite{dean2017sample} and \cite{tu2017least}, and (2) the power system studied in \cite{lewis2012optimal}, Example 11.5-1, with noise $W=I$. 
We start all experiments from an all-zero initial state $x_0 = 0$, and set the initial stable policy $K_1$ to the optimal controller for a system with a modified cost $M' = 200 M$. For simplicity we set $\xi=0$ and $T_s=10$ for \mv{v1}. We set the exploration covariance to $\Sigma_a = I$ for (1) and $\Sigma_a = 10I$ for (2).

In addition to the described algorithms, we also evaluate \mv{v3}, an algorithm identical to \mv{v2} except that the generated datasets $\cZ$ include all data, not just random actions. 
We compare \mv{} to the following:
\begin{itemize}
\item Least squares policy iteration (LSPI) where the policy $\pi_i$ in phase $i$ is greedy with respect to the most recent value function estimate $\widehat Q_{i - 1}$. We use the same estimation procedure as for \mv{}.  
\item A version RLSVI \cite{Osband-etal-2017} where we randomize the value function parameters rather than taking random actions. In particular, we update the mean $\mu_Q$ and covariance $\Sigma_Q$ of a TD estimate of $G$ after each step, and switch to a policy greedy w.r.t. a parameter sample $\widehat{G} \sim (\mu_Q, 0.2 \Sigma_Q)$ every $T^{1/2}$ steps. We project the sample onto the constraint $G \succ \big(\begin{smallmatrix}M & 0 \\ 0 & N \end{smallmatrix}\big)$.
\item A model-based approach which estimates the dynamics parameters $(\widehat{A}, \widehat{B})$ using ordinary least squares. The policy at the end of each phase is produced by treating the estimate as the true parameters (this approach is called \emph{certainty equivalence} in optimal control). 
We use the same strategy as in the model-free case, i.e. we execute the policy for some number of iterations, followed by running the policy and taking random actions.  
\end{itemize} 

To evaluate stability, we run each algorithm 100 times and compute the fraction of times it produces stable policies in all phases.  Figure~\ref{fig:experiments} (left) shows the results as a function of trajectory length.  \mv{v3} is the most stable among model-free algorithms, with performance comparable to the model-based approach.

We evaluate solution cost by running each algorithm until we obtain 100 stable trajectories (if possible), where each trajectory is of length 50,000. We compute both the average cost incurred during each phase $i$, and true expected cost of each policy $\pi_i$. The average cost 
at the end of each phase is shown in Figure~\ref{fig:experiments} (center and right). Overall, \mv{v2} and \mv{v3} achieve lower costs than \mv{v1}, and the performance of \mv{v1} and LSPI is comparable. The lowest cost is achieve by the model-based approach. These results are consistent with the empirical findings of \cite{tu2017least}, where model-based approaches outperform discounted LSTDQ. 

%% file: appendix.tex
\newpage
\appendix
\aistatsfalse
\input{appendix_estimation}

\section{Analysis of the MFLQ algorithm}

\subsection{Proof of Lemma~\ref{lem:greedy-stable}}
\label{sec:stable-proof}
\begin{proof}
Let $G^j = \frac{1}{j} \sum_{i=1}^j G_i$ and $\widehat{G}^j= \frac{1}{j} \sum_{i=1}^j \widehat{G}_i$ be the averages of true and estimated state-action value matrices of policies $K_1, \ldots, K_j$, respectively.  Let $H^j$ and $\widehat{H}^j$ be the corresponding value matrices. 
The greedy policy with respect to $\widehat{G}^j$ is given by:
\begin{align}
K_{j+1} & = \arg \min_K \trace\left( x^\top \begin{bmatrix}I & -K^\top \end{bmatrix} \widehat{G}^j \begin{bmatrix}I \\ -K \end{bmatrix}x \right) \notag \\ & = \arg \min_K \trace \left( \widehat G^j X_K \right) \,, \\
{\rm where}\; & X_K = \begin{bmatrix}I \\ -K \end{bmatrix} xx^\top \begin{bmatrix}I & -K^\top \end{bmatrix}.
\end{align}

Let $|X_K|$ be the matrix obtained from $X_K$ by taking the absolute value of each entry. We have the following:
\begin{align}
\trace(G_j X_{K_{j+1}}) & \leq \trace(\widehat{G}_j X_{K_{j+1}}) + \varepsilon_1 \trace({\bf 1}{\bf 1}^\top |X_{K_{j+1}}|) \label{eq:err1}\\
&\leq \trace(\widehat{G}_j X_{K_j}) + \varepsilon_1 {\bf 1}^\top |X_{K_{j+1}}|{\bf 1} \label{eq:subopt}\\
& \leq \trace(G_j X_{K_j}) + \varepsilon_1 {\bf 1}^\top (|X_{K_j}| + |X_{K_{j+1}}|){\bf 1} \label{eq:err2} \\
&= x^\top H_jx  + \varepsilon_1 {\bf 1}^\top (|X_{K_j}| + |X_{K_{j+1}}|){\bf 1} \label{eq:gh}
\end{align}
Here, \eqref{eq:err1} and \eqref{eq:err2} follow from the error bound,\footnote{Note that the elementwise max norm of a matrix satisfies $\norm{G}_{\max} \leq \norm{G}_F$.} and \eqref{eq:gh} follows from $\trace(G_j X_{K_j}) = x^\top H_j x$. 
 To see \eqref{eq:subopt}, note that $K_{i+1}$ is optimal for $\widehat G^i$ and we have:
\begin{align*}
\trace (\widehat G^j X_{K_{j+1}}) &=  \frac{j-1}{j}  \trace (\widehat G^{j-1} X_{K_{j+1}}) + \frac{1}{j} \trace( \widehat G_j X_{K_{j+1}}) \\
& \leq  \frac{j-1}{j} \trace( \widehat G^{j-1} X_{K_j}) + \frac{1}{j} \trace (\widehat G_j X_{K_j}) \\
& = \trace (\widehat G^j X_{K_{j}}) \,. 
\end{align*}
Since $\trace ( \widehat G^{j-1} X_{K_j}) \leq  \trace( \widehat G^{j-1} X_{K_{j+1}})$ it follows that $\trace (\widehat G_j X_{K_{j+1}}) \leq \trace(\widehat G_j X_{K_j})$. 

Now note that we can rewrite $\trace(G_j X_{K_{j+1}})$ as a function of $H_j$ as follows:
\ifaistats
\begin{align*}
\trace(G_j X_{K_{j+1}}) &= x^\top \begin{bmatrix}I & -K_{j+1}^\top \end{bmatrix}G_j \begin{bmatrix}I \\ -K_{j+1} \end{bmatrix} x  \\
&= x^\top  \begin{bmatrix}I & -K_{j+1}^\top \end{bmatrix} 
\bigg( \begin{bmatrix}A^\top \\ B^\top \end{bmatrix} H_j  \begin{bmatrix} A & B \end{bmatrix} \\
&\qquad \qquad + \begin{bmatrix} M & 0 \\ 0 & N \end{bmatrix} \bigg)\begin{bmatrix}I \\ -K_{j+1} \end{bmatrix} x \\
&= x^\top \bigg( (A - BK_{j+1})^\top H_j (A - BK_{j+1}) \bigg)x  \\
& \quad\quad+ \trace \bigg(\begin{bmatrix} M & 0 \\ 0 & N \end{bmatrix} X_{K_{j+1}}\bigg)  \,.
\end{align*}
\else
\begin{align*}
\trace(G_j X_{K_{j+1}}) &= x^\top \begin{bmatrix}I & -K_{j+1}^\top \end{bmatrix}G_j \begin{bmatrix}I \\ -K_{j+1} \end{bmatrix} x  \\
&= x^\top  \begin{bmatrix}I & -K_{j+1}^\top \end{bmatrix} 
\bigg( \begin{bmatrix}A^\top \\ B^\top \end{bmatrix} H_j  \begin{bmatrix} A & B \end{bmatrix}
+ \begin{bmatrix} M & 0 \\ 0 & N \end{bmatrix} \bigg)\begin{bmatrix}I \\ -K_{j+1} \end{bmatrix} x \\
&= x^\top \bigg( (A - BK_{j+1})^\top H_j (A - BK_{j+1}) \bigg)x + \trace \bigg(\begin{bmatrix} M & 0 \\ 0 & N \end{bmatrix} X_{K_{j+1}}\bigg)  \,.
\end{align*}
\fi
Letting $\Gamma_j = A - BK_j$, we have that 
\begin{align}
\label{eq:stab_ineq2}
x^\top \bigg( \Gamma_{j+1}^\top H_j \Gamma_{j+1} \bigg)x  +  \varepsilon_2 \leq x^\top H_j x
\end{align}
\ifaistats
\begin{align*}
{\rm where} \;\;\varepsilon_2 &= 
x^\top (M + K_{j+1}^\top N K_{j+1})x \\ & \qquad - \varepsilon_1 {\bf 1}^\top (|X_{K_j}| + |X_{K_{j+1}}|){\bf 1}.
\end{align*}
\else
\begin{align*}
{\rm where} \;\;\varepsilon_2 &= 
x^\top (M + K_{j+1}^\top N K_{j+1})x - \varepsilon_1 {\bf 1}^\top (|X_{K_j}| + |X_{K_{j+1}}|){\bf 1}.
\end{align*}
\fi
If the estimation error $\varepsilon_1$ is small enough so that $\varepsilon_2 > 0$ for any unit-norm $x$ and all policies, then $H_j \succ \Gamma_{j+1}^\top H_j \Gamma_{j+1}$ and $K_{j+1}$ is stable by a Lyapunov theorem. Since $K_1$ is stable and $H_1$ bounded, all policies remain stable. 

In order to have $\varepsilon_2 > 0$, it suffices to have 
\begin{equation*}
\varepsilon_1 < \big((\sqrt{n} + \|K_j\|\sqrt{d})^2 + (\sqrt{n} + \|K_{j+1}\|\sqrt{d})^2 \big)^{-1}.
\end{equation*}
This follows since  $M \succ I$, and since for any unit norm vector $x \in \mathbb S^n$, ${\bf 1}^\top x x^\top {\bf 1} \leq n$, with equality achieved by $x = \frac{1}{\sqrt{n}} {\bf 1}$. Similarly, ${\bf 1}^\top Kxx^\top K^\top {\bf 1} \leq \|K\|^2 d$, and ${\bf 1}^\top (|X_{K_j}|) {\bf 1} \leq (\sqrt{n} + \|K_j\| \sqrt{d})^2$.

As we will see, we need a smaller estimation error in phase $j$:
\begin{equation}
\label{eq:small-est-cond}
\varepsilon_1 < \frac{1}{6C_1S} \big((\sqrt{n} + \|K_j\|\sqrt{d})^2 + (\sqrt{n} + \|K_{j+1}\|\sqrt{d})^2 \big)^{-1}.
\end{equation}
Here, $C_1$ is an upper bound on $\norm{H_1}$; note that $H_1 \succ M \succ I$, so $C_1 > 1$. The above condition guarantees that 
\[
\varepsilon_1 {\bf 1}^\top (|X_{K_j}| + |X_{K_{j+1}}|){\bf 1} \le \frac{1}{6C_1S} \;.
\]


We have that $G_{1,22} \succ N \succ I$ and $G_{1,21} = B^\top H_1 A$. Given that the estimation error~\eqref{eq:eps1} is small, we have $\norm{K_2} \le 2( \norm{B^\top H_1 A} + 1) \le C_K$. Then \eqref{eq:eps1} implies \eqref{eq:small-est-cond} for $j=1$, and the above argument shows that $K_2$ is stable. 

Next, we show a bound on $\norm{\Gamma_i^k}$.  Let  $L_{i+1} = H_i^{1/2} \Gamma_{i+1} H_i^{-1/2}$.
By \eqref{eq:stab_ineq2}, $M \succ I$, and the error bound,
\begin{align*}
H_1 &\succ \Gamma_2^\top H_1 \Gamma_2 + (M + K_2^\top N K_2) - (6C_1S)^{-1} I \\
I &\succ L_2^\top L_2 + H_1^{-1/2}(M + K_{2}^\top N K_{2})H_1^{-1/2} - (6C_1S)^{-1} H_1^{-1} \\ 
&\succ L_2^\top L_2 + H_1^{-1} - (6C_1)^{-1} I   \\
&\succ L_2^\top L_2 +  (3C_1)^{-1}I - (6C_1)^{-1}  I  \;.
\end{align*}
Thus, $\norm{L_2} \le \sqrt{1-(6C_1)^{-1}}$ and we have that
\[
\norm{\Gamma_2^k} = \norm{(H_1^{-1/2} L_2 H_1^{1/2})^k} \leq \sqrt{C_1}(1 - (6C_1)^{-1})^{k/2} .\] 
To show a uniform bound on value functions, we first note that 
\[
H_2 - H_1  \prec \Gamma_2^\top (H_2 - H_1) \Gamma_2 + (6C_1S)^{-1} I \;.
\]
Using the stability of $\Gamma_2$, 
\begin{align*}
H_2 - H_1 &\prec (6C_1S)^{-1} \sum_{k=0}^{\infty}(\Gamma_2^\top)^k \Gamma_2^k  \\
\norm{H_2} & \leq \norm{H_1} + \frac{C_1}{6C_1S(1 - \norm{L_2}^2)} \leq (1 + S^{-1}) C_1 \;.
\end{align*}
Thus $C_2 \le (1+S^{-1})C_1$, and by repeating the same argument,
\begin{equation}
C_i \le (1+ S^{-1})^i C_1 \le 3 C_1 \;.
\end{equation}

\end{proof}

\section{Regret bound}
\label{sec:regret_bound}
In this section, we prove Lemma~\ref{lem:bound-beta} by bounding $\beta_T$, $\gamma_T$, and $\alpha_T$.

\subsection{Bounding $\beta_T$}
\label{sec:beta_bound}

Because we use FTL as our expert algorithm and value functions are quadratic, we can use the following regret bound for the FTL algorithm (Theorem 3.1 in \citep{Cesa-Bianchi-Lugosi-2006}).
\begin{theorem}[FTL Regret Bound]
\label{thm:ftl}
Assume that the loss function $f_t(\cdot)$ is convex, is Lipschitz with constant $F_1$, and is twice differentiable everywhere with Hessian $H \succ F_2 I$. Then the regret of the Follow The Leader algorithm is bounded by 
\[
B_T \le \frac{F_1^2}{2 F_2} (1+\log T) \;.
\]
\end{theorem}
Because we execute $S$ policies, each for $\tau = T / S$ rounds (where $\tau = T^{2/3 + \xi}$ and $\tau = T^{3/4}$ for \mv{v1} and \mv{v2}, respectively),
\begin{align*}
\beta_T &= \sum_{i=1}^S \tau \E_{x\sim \mu_{\pi}} ( Q_{i}(x,\pi_i(x)) - Q_{i}(x, \pi(x)) ) \\
&= \tau \sum_{i=1}^S \bigg( \E_{x\sim \mu_{\pi}} ( \widehat Q_{i}(x,\pi_i(x)) - \widehat Q_{i}(x, \pi(x)) )\\ 
&\qquad \qquad  +  \E_{x\sim \mu_{\pi}} ( Q_{i}(x,\pi_i(x)) - \widehat Q_{i}(x, \pi_i(x)) ) \\
&\qquad \qquad  +  \E_{x\sim \mu_{\pi}} ( \widehat Q_{i}(x,\pi(x)) - Q_{i}(x, \pi(x)) ) \bigg) \\
&\le C' \sqrt{ST} \log T + \tau \sum_{i=1}^S \E_{x\sim \mu_{\pi}} ( \widehat Q_{i}(x,\pi_i(x)) - \widehat Q_{i}(x, \pi(x)) ) \,,
\end{align*}
where the last inequality holds by Lemma~\ref{lem:est-error}. Consider the remaining term:
\[
E_T = \tau \sum_{i=1}^S \E_{x\sim \mu_{\pi}} ( \widehat Q_{i}(x,\pi_i(x)) - \widehat Q_{i}(x, \pi(x)) ) \;.
\]
We bound this term using the FTL regret bound. We show that the conditions of Theorem~\ref{thm:ftl} hold for the loss function $ f_i(K) = \E_{x \sim \mu_{\pi}}( \widehat Q_{i}(x, Kx))$. Let $\Sigma_{\pi}$ be the covariance matrix of the steady-state distribution $\mu_\pi(x)$.  We have that
\ifaistats
\begin{align*}
f_i(K) &= \trace\bigg(\Sigma_{\pi} \big( \widehat G_{i, 11}  -  K^\top \widehat G_{i, 21}  - \widehat G_{i, 12}K \\
& \qquad \qquad  + K^\top \widehat G_{i, 22} K \big)\bigg) \\
\nabla_K  f_i(K) &=  2 \Sigma_{\pi} \big( K^\top \widehat G_{i, 22} - \widehat G_{i, 12} \big) \\
&= 2 \mat \big( (\widehat G_{i, 22} \otimes \Sigma_{\pi}) \vect(K) \big) - 2 \Sigma_\pi \widehat G_{i, 12} \\
{\nabla^2_{\vect(K)}} f_i(K) &= 2 \widehat G_{i, 22} \otimes \Sigma_{\pi} \,.
\end{align*}
\else
\begin{align*}
f_i(K) &= \trace\bigg(\Sigma_{\pi} \big( \widehat G_{i, 11}  -  K^\top \widehat G_{i, 21} - \widehat G_{i, 12}K + K^\top \widehat G_{i, 22} K \big)\bigg) \\
\nabla_K  f_i(K) &=  2 \Sigma_{\pi} \big( K^\top \widehat G_{i, 22} - \widehat G_{i, 12} \big) \\
&= 2 \mat \big( (\widehat G_{i, 22} \otimes \Sigma_{\pi}) \vect(K) \big) - 2 \Sigma_\pi \widehat G_{i, 12} \\
{\nabla^2_{\vect(K)}} f_i(K) &= 2 \widehat G_{i, 22} \otimes \Sigma_{\pi} \,.
\end{align*}
\fi
Boundedness and Lipschitzness of the loss function $f_i(K_i)$ follow from the boundedness of policies $K_i$ and value matrix estimates $\widehat G_i$. By Lemma~\ref{lem:greedy-stable}, we have that $\norm{K_i} \leq C_K$. To bound $\| \widehat G_i \|$, note that 
\begin{align*}
G_i &= 
\begin{pmatrix}
           A^\top \\
           B^\top
\end{pmatrix}
H_i
\begin{pmatrix}
          A\,\,\, B
\end{pmatrix}
+
\begin{pmatrix}
  M & 0 \\
  0 & N 
 \end{pmatrix}\; \\
 \norm{G_i} & \le C_H (\norm{A} + \norm{B})^2 + \norm{M} + \norm{N} &{\rm (Lemma~\ref{lem:greedy-stable})}\\
 \norm{\widehat G_i} & \le \norm{G_i} + \varepsilon_1 \sqrt{n + d} & {\rm (Lemma~\ref{lem:est-error})}.
\end{align*} 

The Hessian lower bound is ${\nabla^2_{\vect(K)}} f_i(K) \succ F_2 I$, where $F_2$ is given by two times the product of the minimum eigenvalues of $\Sigma_\pi$ and $\widehat G_{i, 22}$. For any stable policy $\pi(x) = Kx$, the covariance matrix of the stationary distribution satisfies $\Sigma_\pi \succ W$, and we project the estimates $\widehat G_i$ onto the constraint 
$\widehat G_i \succeq (\begin{smallmatrix} M & 0 \\ 0 & N \end{smallmatrix})$. Therefore the Hessian of the loss is lower-bounded by $2\lambda_{\min}(W) I$. 
By Theorem~\ref{thm:ftl}, $E_T \le \tau \log S = C'' \tau \log T$ for an appropriate constant $C''$. 


\subsection{Bounding $\gamma_T$}
\label{sec:gamma_bound}

In this section, we bound the average cost of following a stable policy, $\gamma_T = \sum_{t=1}^T (\lambda_{\pi} -  c(x_t, \pi(x_t)))$.
Recall that the instantaneous  and average costs of following a policy $\pi(x) = -Kx$ can be written as
\begin{align}
 c(x_t, \pi(x_t)) &= x_t^\top (M + K^\top N K) x_t \\
 \lambda_\pi &= \trace(\Sigma_\pi (M + K^\top N K)) \;, 
\end{align}
where $\Sigma_\pi$ is the steady-state covariance of $x_t$. 
Let $\Sigma_t$ be the covariance of $x_t$, let $D_t = \Sigma_t^{1/2}(M + K^\top N K)\Sigma_t^{1/2}$, and let $\lambda_t = \trace(D_t)$. 
To bound $\gamma_T$, we start by rewriting the cost terms as follows:
\ifaistats
\begin{align*}
\lambda_\pi - c(x_t, \pi(x_t)) &= \lambda_\pi - \lambda_t + \lambda_t - c(x_t^\pi, \pi(x_t^\pi)) \\
&= \trace((\Sigma_\pi - \Sigma_t)(M + K^\top N K))   \\
& \qquad +  \trace(D_t) - u_t^\top D_t u_t
\end{align*}
\else
\begin{align}
\lambda_\pi - c(x_t, \pi(x_t)) &= \lambda_\pi - \lambda_t + \lambda_t - c(x_t^\pi, \pi(x_t^\pi)) \\
&= \trace((\Sigma_\pi - \Sigma_t)(M + K^\top N K))  + \big( \trace(D_t) - u_t^\top D_t u_t\big)
\end{align}
\fi
where $u_t \sim \mathcal{N}(0, I_n)$ is a standard normal vector.

To bound $\trace((\Sigma_\pi - \Sigma_t)(M + K^\top N K))$, note that $\Sigma_\pi = \Gamma \Sigma_{\pi} \Gamma^\top + W$ and $\Sigma_t = \Gamma \Sigma_{t-1}  \Gamma^\top + W$.  Subtracting the two equations and recursing, 
\begin{align}
\Sigma_\pi - \Sigma_t =  \Gamma (\Sigma_\pi - \Sigma_{t-1}) \Gamma^\top =  \Gamma^t(\Sigma_\pi - \Sigma_0)(\Gamma^t)^\top.
\end{align}
Thus, 
\ifaistats
\begin{align}
& \sum_{t=0}^T \trace((M + K^\top N K) (\Sigma_\pi - \Sigma_t))  \notag \\
&\qquad =  \sum_{t=0}^T  \trace \big( (M + K^\top N K) \Gamma^t(\Sigma_\pi - \Sigma_0)(\Gamma^t)^\top \big) \notag \\
&\qquad \leq \trace(\Sigma_\pi - \Sigma_0) \trace \bigg( \sum_{t=0}^\infty (\Gamma^t)^\top (M + K^\top N K) \Gamma^t \bigg) \notag \\
& \qquad = \trace(\Sigma_\pi - \Sigma_0) \trace( H_\pi ) \;.
\label{eq:sigma_bound}
\end{align}
\else
\begin{align}
\sum_{t=0}^T \trace((M + K^\top N K) (\Sigma_\pi - \Sigma_t)) 
&=  \sum_{t=0}^T  \trace \big( (M + K^\top N K) \Gamma^t(\Sigma_\pi - \Sigma_0)(\Gamma^t)^\top \big) \\
&\leq \trace(\Sigma_\pi - \Sigma_0) \trace \bigg( \sum_{t=0}^\infty (\Gamma^t)^\top (M + K^\top N K) \Gamma^t \bigg)\\
&= \trace(\Sigma_\pi - \Sigma_0) \trace( H_\pi ) \;.
\label{eq:sigma_bound}
\end{align}
\fi

Let $U$ be the concatenation of $u_1,\dots,u_T$, and let $D$ be a block-diagonal matrix constructed from $D_1,\dots,D_T$. To bound the second term, note that by the Hanson-Wright inequality 
\ifaistats
\begin{align}
\notag
& \Prob{\abs{\sum_{t=1}^T u_t^\top D_t u_t - \trace{D_t}} > s} \\
&\qquad = \Prob{\abs{U^\top D U - \trace{D}} > s } \notag \\
&\qquad \leq 2 \exp\bigg(-c \min\bigg(\frac{s^2}{\norm{D}_F^2}, \frac{s}{\norm{D}}\bigg)\bigg)
\end{align}
\else
\begin{align}
\notag
\Prob{\abs{\sum_{t=1}^T u_t^\top D_t u_t - \trace{D_t}} > s} &= \Prob{\abs{U^\top D U - \trace{D}} > s } \notag \\
&\leq 2 \exp\bigg(-c \min\bigg(\frac{s^2}{\norm{D}_F^2}, \frac{s}{\norm{D}}\bigg)\bigg) \,.
\end{align}
\fi
Thus with probability at least $1-\delta$ we have
\ifaistats
\begin{align}
\notag
& \abs{\sum_{t=1}^T u_t^\top D_t u_t - \trace(D_t) } \\
&\qquad \leq \norm{D}_F \sqrt{\ln(2/\delta) / c} + \norm{D} \ln(2/\delta)/c \notag \\
&\qquad \le \sqrt{\sum_{t=1}^T \norm{D_t}_F^2} \sqrt{\ln(2/\delta) / c} + \max_t \norm{D_t} \ln(2/\delta)/c
\end{align}
\else
\begin{align}
\notag
\abs{\sum_{t=1}^T u_t^\top D_t u_t - \trace(D_t) } &\leq \norm{D}_F \sqrt{\ln(2/\delta) / c} + \norm{D} \ln(2/\delta)/c\\
&\le \sqrt{\sum_{t=1}^T \norm{D_t}_F^2} \sqrt{\ln(2/\delta) / c} + \max_t \norm{D_t} \ln(2/\delta)/c
\end{align}
\fi
where $c$ is a universal constant. 
Given that for all $t$, 
\begin{align*}
\norm{D_t} & \leq \trace(D_t) \\ & = \trace((M + K^\top N K)(\Sigma_\pi + \Gamma^t(\Sigma_0 - \Sigma_\pi)(\Gamma^T)^t)) \\ & \leq \lambda_\pi\,,
\end{align*}
with probability at least $1-\delta$,
\beq
\sum_{t=1}^T c(x_t^\pi, \pi(x_t^\pi)) - \lambda_t \leq  \lambda_\pi  \left( \sqrt{T \ln(2/\delta) / c} + \ln(2/\delta)/c \right) \;.
\eeq
Thus, we can bound $\gamma_T$ as
\beq
\gamma_T  \leq \trace(H_\pi)\trace(\Sigma_\pi) + \lambda_\pi  \left( \sqrt{T \ln(2/\delta) / c} + \ln(2/\delta)/c \right) \;.
\eeq

\subsection{Bounding $\alpha_T$}
\label{sec:alpha_bound}

To bound  $\alpha_T = \sum_{t=1}^T (c(x_t,a_t) - \lambda_{\pi_t})$, in addition to bounding the cost of following a policy, we need to account for having $S$ policy switches, as well as the cost of random actions. 
Let $I_a$ be the set of time indices of all random actions $a \sim \cN(0, \Sigma_a)$. Using the Hanson-Wright inequality, with probability at least $1-\delta$,
\ifaistats
\begin{align*}
\sum_{t \in I_a} |a_t^\top N a_t - \trace(\Sigma_a N)| \leq & \norm{\Sigma_a N}_F \sqrt{|I_a|\ln(2/\delta) / c_1} \\ & + \norm{\Sigma_a N} \ln(2/\delta)/c_1 \;.
\end{align*}
\else
\beq
\sum_{t \in I_a} |a_t^\top N a_t - \trace(\Sigma_a N)| \leq \norm{\Sigma_a N}_F \sqrt{|I_a|\ln(2/\delta) / c_1} + \norm{\Sigma_a N} \ln(2/\delta)/c_1 \;.
\eeq
\fi
Let $D_{i,t} = \Sigma_{t}^{1/2}(M + K_i^\top N K_i) \Sigma_t^{1/2}$, and let $\lambda_{i, t} = \trace(D_{i, t})$. Let $I_{i}$ be the set of time indices corresponding to following policy $\pi_i$ in phase $i$. The corresponding cost can be decomposed similarly to $\gamma_T$:
\ifaistats
\begin{align}
\label{eq:alpha}
& \sum_{i=1}^{S} \sum_{t \in I_{i}} c(x_t, \pi_i(x_t)) - \lambda_{\pi_i} \notag \\
&\qquad = \sum_{i=1}^{S} \sum_{t \in I_{i}} \bigg( \trace((\Sigma_t - \Sigma_{\pi_i})(M + K_i^\top N K_i) \notag \\
&\qquad\qquad\qquad +  u_t^\top D_{i,t} u_t - \trace(D_{i, t}) \bigg) \;.
\end{align}
\else
\begin{align}
\label{eq:alpha}
\sum_{i=1}^{S} \sum_{t \in I_{i}} c(x_t, \pi_i(x_t)) - \lambda_{\pi_i} &= \sum_{i=1}^{S} \sum_{t \in I_{i}}  \trace((\Sigma_t - \Sigma_{\pi_i})(M + K_i^\top N K_i) + \big( u_t^\top D_{i,t} u_t - \trace(D_{i, t}) \big) \;.
\end{align}
\fi
Let $D_{\max} \geq \max_{i, t} \norm{D_{i,t}}$. Similarly to the previous section, with probability at least $1-\delta$ we have
\ifaistats
\begin{align*}
\abs{\sum_{i=1}^{S} \sum_{t \in I_{i}} u_t^\top D_{i,t} u_t - \trace(D_{i, t})  } 
&\leq D_{\max} \sqrt{T n\ln(2/\delta) / c_2}  \\&\quad+ D_{\max} \ln(2/\delta)/c_2 \;.
\end{align*}
\else
\begin{align}
\abs{\sum_{i=1}^{S} \sum_{t \in I_{i}} u_t^\top D_{i,t} u_t - \trace(D_{i, t})  } 
&\leq D_{\max} \sqrt{T n\ln(2/\delta) / c_2} + D_{\max} \ln(2/\delta)/c_2 \;.
\end{align}
\fi
At the beginning of each phase $i$, the state covariance is $\Sigma_{\pi_{i-1}}$ (and we define $\Sigma_{\pi_0}=W$). After following $\pi_i$ for $T_v$ steps,  
\ifaistats
\begin{align*}
& \sum_{i=1}^S \sum_{t \in I_i} \trace((\Sigma_t - \Sigma_{\pi_i})(M + K_i^\top N K_i)) \\
&\qquad = \sum_{i=1}^S \sum_{k =0}^{T_v - 1} \trace(  (\Sigma_{\pi_{i-1}} - \Sigma_{\pi_i}) ({\Gamma_i}^k)^\top (M + K_i^\top NK_i) {\Gamma_i}^k)   \\
& \qquad \leq \sum_{i=1}^S \trace(H_i) \trace(\Sigma_{\pi_{i-1}}) \\
 & \qquad \leq  S n C_H   \max_i \trace(\Sigma_{\pi_i})
\end{align*}
\else
\begin{align*}
\sum_{i=1}^S \sum_{t \in I_i} \trace((\Sigma_t - \Sigma_{\pi_i})(M + K_i^\top N K_i))
&= \sum_{i=1}^S \sum_{k =0}^{T_v - 1} \trace(  (\Sigma_{\pi_{i-1}} - \Sigma_{\pi_i}) ({\Gamma_i}^k)^\top (M + K_i^\top NK_i) {\Gamma_i}^k)   \\
& \leq \sum_{i=1}^S \trace(H_i) \trace(\Sigma_{\pi_{i-1}}) \\
 & \leq  S n C_H   \max_i \trace(\Sigma_{\pi_i})
\end{align*}
\fi
Following each random action, the state covariance is $\Sigma_{G,i} = A \Sigma_{\pi_i} A^\top +  B\Sigma_a B^\top + W$. After taking a random action and following $\pi_i$ for $T_s$ steps, we have 
\ifaistats
\begin{align*}
& \sum_{k=0}^{T_s} \trace((\Sigma_{G, i} - \Sigma_{\pi_i}) ({\Gamma_i}^k)^\top (M + K_i^\top NK_i) {\Gamma_i}^k) \\
&\qquad \leq  \trace(\Sigma_{G, i}) \trace(H_i)  \\
 &\qquad \leq n C_H (\trace(B \Sigma_a B^\top) + \norm{A}^2 \trace(\Sigma_{\pi_i})) \;.
\end{align*}
\else
\begin{align*}
\sum_{k=0}^{T_s} \trace((\Sigma_{G, i} - \Sigma_{\pi_i}) ({\Gamma_i}^k)^\top (M + K_i^\top NK_i) {\Gamma_i}^k) & \leq  \trace(\Sigma_{G, i}) \trace(H_i) 
 \leq n C_H (\trace(B \Sigma_a B^\top) + \norm{A}^2 \trace(\Sigma_{\pi_i})) \;.
\end{align*}
\fi
Putting everything together, we have
\begin{align*}
\alpha_t 
& \leq \norm{\Sigma_a N}_F \sqrt{|I_a|\ln(2/\delta) / c_1} + \norm{\Sigma_a N} \ln(2/\delta)/c_1  \notag \\
& + D_{\max} \sqrt{T n\ln(2/\delta) / c_2} + D_{\max} \ln(2/\delta)/c_2 \\
& + S nC_H  \max_i \trace(\Sigma_{\pi_{i}}) \\
& + |I_a| n C_H \big( \trace(B \Sigma_a B^\top) + \norm{A}^2 \max_i \trace(\Sigma_{\pi_i}) \big) 
\end{align*}
where in \ver{1} $S = T^{1/3 - \xi}$ and $|I_a| = O(T^{2/3 + \xi})$, while in \ver{2} $S = T^{1/4}$ and $|I_a| = T^{3/4 + \xi}$. We bound $\max_i \trace(\Sigma_{\pi_i})$ and $\norm{D_{i,t}}$ in \ref{sec:cov_bound}.

\subsubsection{State covariance bound}
\label{sec:cov_bound}

We bound $\max_i \trace(\Sigma_{\pi_i})$ using the following equation for the average cost of a policy:
 \begin{align*}
\trace(\Sigma_{\pi_i}( M + K_i^\top N K_i)) & = \trace(H_i W) \\
\trace(\Sigma_{\pi_i}) & \leq \norm{H_i} \trace(W) / \lambda_{\min}(M) \\
\max_i \trace(\Sigma_{\pi_i}) & \leq C_H \trace(W) / \lambda_{\min}(M) \;.
 \end{align*}
To bound $\norm{D_{i, t}}$, we note that 
\begin{align*}
\norm{D_{i,t}} \leq \trace(D_{i,t}) &= \trace\big(\Sigma_t (M + K_i^\top N K_i)\big) \\
& \leq \trace(\Sigma_t) (\norm{M} + C_K^2 \norm{N})\;,
\end{align*}
and bound the state covariance $\trace(\Sigma_t)$. After starting at distribution $\cN(0, \Sigma_0)$ and following a policy $\pi_i$ for $t$ steps, the state covariance is 
\begin{align*}
\label{eq:cov_bound}
\Sigma_{t} &= \Gamma_i \Sigma_{t-1} \Gamma_i^\top + W \\
 &= \Gamma_i^t \Sigma_0 {\Gamma_i^t}^{\top} + \sum_{k=0}^{t-1} \Gamma_i^k W {\Gamma_i^k}^\top  \\
 & \prec \Sigma_0 + \Sigma_{\pi_i} \;.
\end{align*}
The initial covariance $\Sigma_0$ is close to $\Sigma_{\pi_{i-1}}$ after a policy switch, and close to $A \Sigma_{\pi_i} A^\top + B \Sigma_a B^\top + W$ after a random action.  Therefore we can bound the state covariance in each phase as
\begin{align*}
\Sigma_t &\preceq \Sigma_{\pi_i} +  \Sigma_{\pi_{i-1}} + A \Sigma_{\pi_i} A^\top + B \Sigma_a B^\top \\
\trace(\Sigma_t) &\leq (2 + \norm{A}^2) \max_i \trace(\Sigma_{\pi_i}) + \trace(B \Sigma_a B^\top)\\
&\leq (2 + \norm{A}^2) C_H \trace(W) / \lambda_{\min}(M) + \trace(B \Sigma_a B^\top) \;.
\end{align*}

%% file: appendix_estimation.tex

\section{Value function estimation}

\subsection{Proof of Lemma~\ref{lem:estv}}
\label{app:estv_proof}

To see the identity 
\begin{align}
P_{\Phi}(\Phi - \overline{\Phi}_+ + {\bf W})(h - \widehat{h}) &= \Phi (I - \Gamma \otimes \Gamma)^\top \vect(h - \hat h),
\end{align}
note that a single element of the vector $(\Phi - \overline{\Phi}_+ + W)(h - \widehat{h})$ can be expressed as
\ifaistats
\begin{align}
& (\phi - \E(\phi_+) + \vect(W))^\top (h - \hat{h}) \notag \\
&\;\;= \vect( xx^\top - \Gamma x x^\top \Gamma^\top)^\top (h - \hat{h})  \notag \\
&\;\;=  \vect(xx^\top)^\top (I - \Gamma \otimes \Gamma)^\top  (h - \hat{h}),
\end{align}
\else
\begin{align}
(\phi - \E(\phi_+) + \vect(W))^\top (h - \hat{h}) 
&= \vect( xx^\top - \Gamma x x^\top \Gamma^\top)^\top (h - \hat{h}) 
=  \vect(xx^\top)^\top (I - \Gamma \otimes \Gamma)^\top  (h - \hat{h}),
\end{align}
\fi
where we have used the Kronecker product identity $\vect(\Gamma X \Gamma^\top ) = (\Gamma \otimes \Gamma) \vect(X)$. Thus we have that
\begin{align}
\norm{\Phi (I - \Gamma \otimes \Gamma)^\top (h - \hat h)} &\leq \norm{P_{\Phi}( \overline{\Phi}_+ - \Phi_+) \hat{h}} \,.
\end{align}
%

Next we lower-bound  $\norm{ (I - \Gamma \otimes \Gamma)^\top  (h - \hat h)}$. Let $L = H^{1/2} \Gamma H^{-1/2}$ and let $\overline H = I - H^{-1/2}\estH H^{-1/2}$. 
We have the following:
\begin{align*}
 \norm{ (I - \Gamma \otimes \Gamma)^\top  \vect(H - \widehat H)} 
& =  \norm{ H - \estH - \Gamma^\top (H - \estH) \Gamma}_F  \\
&=  \norm{ H^{1/2}( \overline{H} - L^\top \overline{H} L)  H^{1/2}}_F \\ 
&= \sqrt{ \trace( H (\overline{H} - L^\top \overline H L) H (\overline{H} - L^\top \overline H L))} \\
& \geq \lambda_{\min}(H)  \norm{ \overline{H} - L^\top \overline{H} L}_F \\
&  \geq \lambda_{\min}(M)  \norm{ \overline{H} - L^\top \overline{H} L}_F \;.
\end{align*}
where the second-last inequality follows from the fact that $\trace(AB) \geq \lambda_{\min}(A) \trace(B)$ for p.s.d matrices $A$ and $B$ \citep{zhang2006eigenvalue}. Furthermore, using the fact that $\norm{L}^2 \leq 1 - \lambda_{\min}(M) \norm{H}^{-1}$, \footnote{This can be seen by multiplying the equation $H  \succ \Gamma^\top H \Gamma + \lambda_{\min}(M)I$ by $H^{-1/2}$ on both sides.}
\begin{align*}
 \norm{ \overline{H} - L^\top \overline{H} L}_F  
&  =  \norm{ (I - L \otimes L)^\top \vect(\overline H)} \\
 &  \geq (1 - \norm{L}^2) \norm{ \overline H}_F \\
&\geq \frac{ \lambda_{\min}(M)}{ \norm{H}} \norm{I - H^{-1/2} \estH H^{-1/2}}_F  \\
& = \frac{ \lambda_{\min}(M)}{ \norm{H}}\sqrt{\trace \big( H^{-1}(H - \estH) H^{-1} (H - \estH) \big)} \\
& \geq \lambda_{\min}(M) \norm{H}^{-2} \norm{H - \estH}_F \;.
\end{align*}
Hence we get that 
\beq
\norm{ (I - \Gamma \otimes \Gamma)^\top  (h - \hat h)} \geq \lambda_{\min}(M)^2 \norm{H}^{-2} \norm{H - \estH}_F\,.
\eeq

\subsection{Proof of Lemma~\ref{lem:est-error}}
\label{app:estg_proof}

\begin{proof}
  Let $P_{\Psi} = \Psi(\Psi^\top\Psi)^{-1}\Psi$ be the orthogonal projector onto $\Psi$.  The true parameters $g = \vect(G)$ and the estimate $\hat{g}=\vect(\estG)$ satisfy the following:
\begin{align}
\Psi \hat{g} &= P_{\Psi}({\bf c} + (\Phi_+ - {\bf W}) \hat{h}) \\
\Psi g &= {\bf c}  + (\overline{\Phi}_+ - {\bf W}) h 
\end{align}
Subtracting the above equations, we have 
\begin{align*}
\norm{\Psi g - \Psi \hat{g}} &= \norm{P_{\Psi} \big( (\overline{\Phi}_+ - {\bf W}) (h - \hat{h}) + (\overline \Phi_+ - \Phi_+) \hat{h}\big)} \\
& \leq \norm{ (\overline{\Phi}_+ - {\bf W}) (h - \hat{h})} + \norm{P_{\Psi}  (\overline \Phi_+ - \Phi_+) \hat{h}}.
\end{align*}
Using $\norm{\Psi v} \geq \sqrt{\lambda_{\min}(\Psi^\top \Psi)}\norm{v}$ on the l.h.s., and $\norm{P_{\Psi} v} \leq \norm{\Psi^\top v} / \sqrt{\lambda_{\min}(\Psi^\top \Psi)}$ on the r.h.s.,
\ifaistats
\begin{align}
& \norm{G - \estG}_F   = \norm{g - \hat{g}} \notag \\
&\qquad \leq \frac{\norm{(\overline{\Phi}_+ - {\bf W}) (h - \hat{h})}} { \sqrt{\lambda_{\min}(\Psi^\top \Psi)}} 
 + 
\frac{ \norm{\Psi^\top(\overline \Phi_+ - \Phi_+) \hat{h}}}{\lambda_{\min}(\Psi^\top \Psi)} .
\label{eq:qerror}
\end{align}
\else
\begin{align}
\norm{G - \estG}_F   = \norm{g - \hat{g}} 
& \leq \frac{\norm{(\overline{\Phi}_+ - {\bf W}) (h - \hat{h})}} { \sqrt{\lambda_{\min}(\Psi^\top \Psi)}} 
 + 
\frac{ \norm{\Psi^\top(\overline \Phi_+ - \Phi_+) \hat{h}}}{\lambda_{\min}(\Psi^\top \Psi)} .
\label{eq:qerror}
\end{align}
\fi
%
Using similar arguments as for $\lambda_{\min}(\Phi^\top \Phi)$ and the fact that actions are randomly sampled, it can be shown that $\lambda_{\min}(\Psi^\top \Psi) = O(\tau)$. 

Let $\Sigma_{G, \pi} = A \Sigma_\pi A^\top + B \Sigma_a B^\top$. 
Assuming that we are close to steady state $x \sim \cN(0, \Sigma_\pi)$ each time we take a random action $a \sim \cN(0, \Sigma_a)$, the next state is distributed as $x_+ \sim \cN(0, \Sigma_{G, \pi} + W)$. Therefore each element of $(\overline{\Phi}_+ - {\bf W}) (h - \hat{h})$ is bounded as:
\begin{align*}
|(\E(\phi_+) - \vect(W))^\top (h - \hat{h})| &= |\trace \big( \Sigma_{G, \pi} (H - \estH)\big)| \notag \\
& \leq \trace( \Sigma_{G, \pi}) \norm{H - \estH} ,
\end{align*}
where we have used the fact that $|\trace(M_1M_2)| \leq \norm{M_1}\trace(M_2)$ for real-valued square matrices $M_1$ and $M_2 \succ 0$  (see e.g. \citep{zhang2006eigenvalue}). Thus, the first term of \eqref{eq:qerror} is bounded as  
\begin{align}
\norm{(\overline{\Phi}_+ - {\bf W}) (h - \hat{h})}
& \leq \trace(\Sigma_{G, \pi})  \norm{H - \widehat H}  \sqrt{\tau}.
\end{align}

To bound the second term, we can again use Lemma 4.8 of \cite{tu2017least}, where the only changes are that we bound $\max_t \norm{\psi_t}$ as opposed to $\max_t \norm{\phi_t}$, and that we have a different distribution of next-state vectors $x_+$.  Thus, with probability at least $1-\delta$, the second term scales as
\ifaistats
\begin{align*}
& \norm{\Psi^\top(\overline \Phi_+ - \Phi_+) \hat{h}} =  \\
& \qquad O \bigg( \sqrt{\tau}  \norm{ W \widehat{H}}_F  (\trace(\Sigma_{\pi}) + \trace(\Sigma_a)) \norm{\Sigma_{G, \pi}}_F  \\
& \quad \quad \quad \quad {\rm polylog}(n^2, 1/\delta, \tau) \bigg).
\end{align*} 
\else
\begin{align}
\norm{\Psi^\top(\overline \Phi_+ - \Phi_+) \hat{h}} = O \bigg( \sqrt{\tau}  \norm{ W \widehat{H}}_F  (\trace(\Sigma_{\pi}) + \trace(\Sigma_a)) \norm{\Sigma_{G, \pi}}_F {\rm polylog}(n^2, 1/\delta, \tau) \bigg).
\end{align} 
\fi
\end{proof}